\documentclass{article}

\usepackage[preprint]{neurips_2026}

\usepackage[utf8]{inputenc}
\usepackage[T1]{fontenc}
\usepackage{float}
\usepackage{algorithm}
\usepackage{algpseudocode}
\usepackage{hyperref}
\usepackage{url}
\usepackage{booktabs}
\usepackage{amsmath}
\usepackage{amssymb}
\usepackage{amsfonts}
\usepackage{microtype}
\usepackage[table]{xcolor}
\usepackage{graphicx}
\usepackage{array}
\usepackage{listings}
\usepackage[most]{tcolorbox}
\usepackage{enumitem}

\newcommand{\method}{Generation Navigator}
\newcommand{\pregrpo}{PRE-GRPO}
\newcommand{\stopact}{\textsc{Stop}}
\newcommand{\refineact}{\textsc{Refine}}
\newcommand{\regenact}{\textsc{Regenerate}}
\lstdefinestyle{promptbox}{
  basicstyle=\ttfamily\tiny,
  breaklines=true,
  breakatwhitespace=false,
  columns=fullflexible,
  keepspaces=true,
  showstringspaces=false,
  frame=single,
  xleftmargin=0.5em,
  xrightmargin=0.5em,
  aboveskip=0.6em,
  belowskip=0.6em
}
\newcommand{\qualcoloredcard}[5]{%
  \begin{minipage}[t]{#1}
  \setlength{\fboxsep}{3pt}\setlength{\fboxrule}{0.7pt}%
  \noindent\fcolorbox{#5}{white}{\begin{minipage}[t]{\dimexpr\linewidth-2\fboxsep-2\fboxrule\relax}
  \centering
  \includegraphics[width=\linewidth]{#2}\\[0.25em]
  {\scriptsize\textbf{#3}\par}
  {\scriptsize #4\par}
  \end{minipage}}
  \end{minipage}}
\newcommand{\qualcard}[4]{\qualcoloredcard{#1}{#2}{#3}{#4}{green!45!black}}
\newcommand{\qualbasecard}[4]{\qualcoloredcard{#1}{#2}{#3}{#4}{red!70!black}}
\newcommand{\qualkey}[1]{\textcolor{red!75!black}{#1}}
\newcommand{\graycell}[1]{\textcolor{black!55}{#1}}

\tcbset{
  qualdetail/.style={
    breakable,
    colback=white,
    colframe=black!35,
    colbacktitle=black!6,
    coltitle=black,
    fonttitle=\bfseries\small,
    boxrule=0.4pt,
    arc=1pt,
    left=5pt,
    right=5pt,
    top=5pt,
    bottom=5pt,
    before skip=0.7em,
    after skip=0.7em
  }
}

\title{Generation Navigator: A State-Aware Agentic Framework for Image Generation}

\author{Jinming Liu$^{1,2,*}$, Ruoyu Feng$^{3,*}$, Yuqi Wang$^{3}$, Wenjun Zeng$^{2}$, Xin Jin$^{2,\dagger}$ \\
$^1$Shanghai Jiao Tong University \quad
    $^2$Eastern Institute of Technology, Ningbo, China \quad
    $^3$Independent}

\begin{document}
\raggedbottom

\maketitle
\renewcommand\thefootnote{\fnsymbol{footnote}}
\footnotetext[1]{Equal contribution.}
\footnotetext[2]{Corresponding author: Xin Jin (jinxin@eitech.edu.cn).}

\begin{abstract}
% Generating images from complex prompts typically requires manual multi-turn trial and error.
Despite rapid advances in text-to-image generation, faithfully realizing user intent remains challenging, often requiring manual multi-turn trial and error.
To automate this process, existing systems rely on either simple prompt rewriting or closed-loop agents driven by hand-crafted rules, rather than learning to adapt actions to the evolving generation process.
In this paper, we reformulate image generation as a state-conditioned action-making problem and propose Generation Navigator, a multi-turn T2I agent that learns to dynamically steer the generation trajectory and output the next action.
However, training this agent via reinforcement learning introduces a critical credit assignment challenge: naively rewarding a trajectory based solely on a single state assigns equal credit to all actions in the rollout, ignores the
quality dynamics across turns, and fails to distinguish actions that improve the trajectory from those that degrade it or waste turns without progress.
We resolve this with PRE-GRPO (Peak-Retention-Efficiency Group Relative Policy Optimization), a trajectory-level reinforcement learning objective that explicitly rewards discovering a high-quality image (Peak), avoiding subsequent quality degradation across turns (Retention), and minimizing unnecessary turns (Efficiency).
Experiments show substantial improvements across benchmarks, reaching a WISE score of 0.90 and 79.06\% reasoning accuracy on T2I-ReasonBench.

\end{abstract}

\section{Introduction}
\label{sec:intro}
Text-to-image (T2I) generation models have made remarkable progress \citep{rombach2022latentdiffusion,saharia2022imagen,esser2024sd3}. Yet satisfying user intent remains challenging in a single pass, especially when the intended image involves spatial layouts, commonsense reasoning, fine-grained style, or ambiguous details.
In practice, users address this through trial and error: generating an image,
inspecting the result, editing it or restarting from scratch depending on its
current quality. This suggests that such T2I generation should not be treated merely as a one-shot prompt-to-image mapping, but as a sequential action-making process: after each generation, choosing the next action given the current result. 

However, existing methods usually model such T2I generation merely as a prompt rewriting or fixed-workflow task. Prompt-centric methods \citep{lian2024llmgrounded,yang2024rpg,wang2025promptenhancer} improve initial prompts but operate only before
generation, leaving no recourse even though the resulting image is partially correct or
structurally flawed.
Recent closed-loop or agentic systems \citep{jiang2026genagent,wan2025maestro} introduce reviewer (e.g., multi-modal large language models) feedback or self-critique, but still rely on fixed workflows or hand-crafted editing rules rather than learning a state-conditioned action policy, which differs from real user behavior.

% To quantify the impact of the gap between fixed workflows and real user behavior, we conduct a pilot study on T2I-ReasonBench. Starting from an initial image produced by a single-pass generator, we construct an empirical action-preference reference. At each turn, the reference executes two possible next actions—\refineact{} (editing the current image) and \regenact{} (producing a new image from a revised prompt)—scores each outcome with the benchmark evaluator, and proceeds along the higher-scoring branch into the next turn.
This raises a natural question: does dynamic state-conditioned action making actually outperform fixed workflows? To answer this, we conduct a pilot study that quantifies the benefit of dynamic action. Starting from an initial image produced by a single-pass generator, we construct an empirical action-preference reference. Mimicking real user behavior-where one either refines an acceptable draft or revises the prompt and retries—the reference at each turn executes two candidate actions, REFINE (editing the current image) and REGENERATE (producing a new image from a revised prompt), scores each outcome with the benchmark evaluator, and proceeds along the higher-scoring branch into the next turn (detailed settings are in the Appendix~\ref{app:pilot_study}).
Results in Fig.~\ref{fig:teaser}(a) show that \regenact{} wins 47.01\% of cases, \refineact{} wins 39.38\%, and 13.61\% are ties. At the system level, fixed refine-only and regenerate-only workflows improve over one-shot generation but still underperform state-conditioned action making (training-free preference reference and our method). These results suggest that \textbf{the optimal action is state-dependent and fixed workflows are suboptimal.}

To learn dynamic state-conditioned actions, reinforcement learning (RL) is a natural tool. However, it raises a second bottleneck: credit assignment. Vanilla RL methods~\citep{shao2024deepseekmath} reward only a single state within the trajectory—typically the final or the best-generated image—while ignoring the quality dynamics across the full generation trajectory. More concretely, rewarding only the final image discards useful mid-trajectory signal, making learning inefficient; rewarding only the best image credits all actions equally, whether they improve, degrade, or waste turns, leading to suboptimal trajectories (Fig.~\ref{fig:teaser}(b)). Thus, \textbf{a good reward should capture the trajectory-level quality: } whether a high-quality image is discovered, whether quality is improved turn-by-turn, and whether the trajectory is efficient without unnecessary turns.

\begin{figure}[tbp]
  \centering
  \begin{minipage}[t]{0.43\linewidth}
    \vspace{0pt}
    \centering
    \vspace*{-0.2em}
    \includegraphics[width=\linewidth]{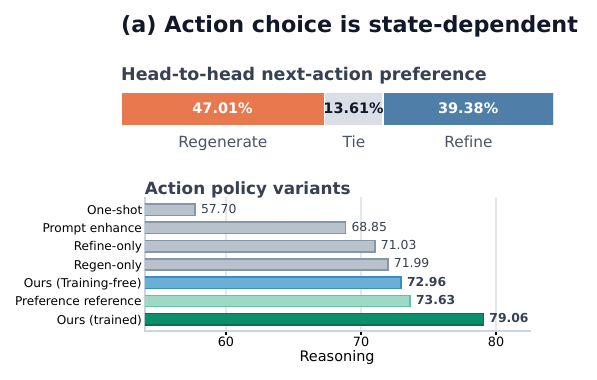}
  \end{minipage}\hfill%
  \begin{minipage}[t]{0.56\linewidth}
    \vspace{0pt}
    \centering
    \includegraphics[width=\linewidth]{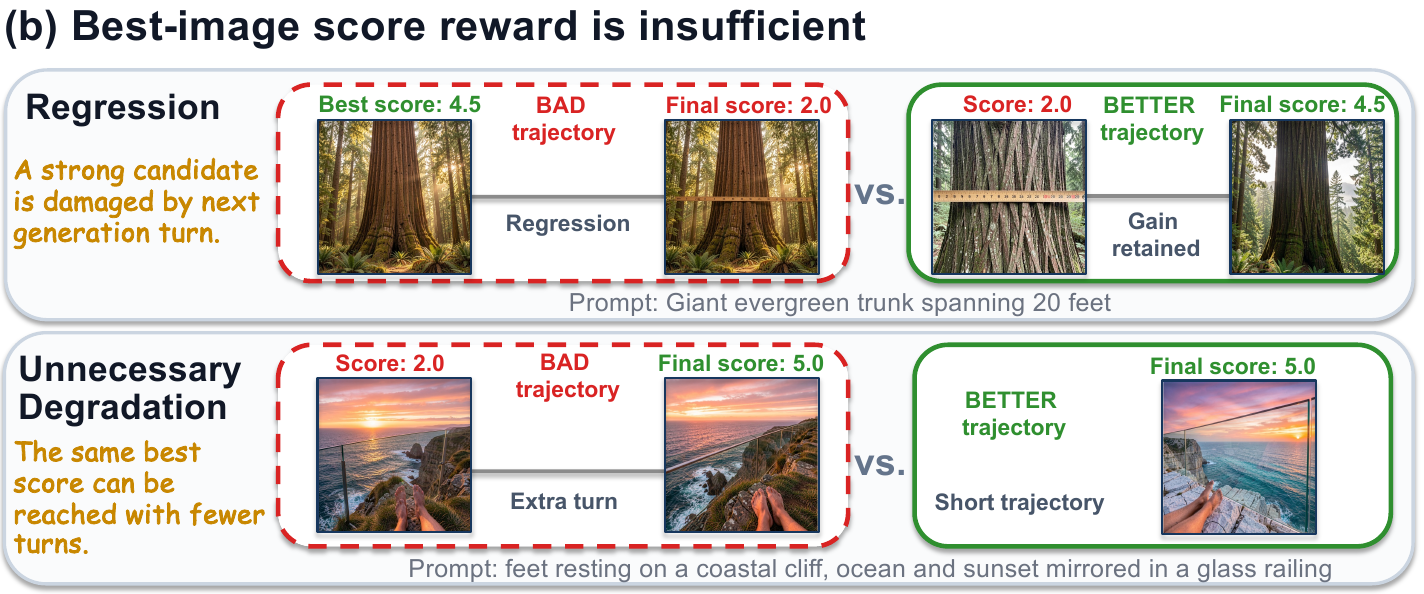}
  \end{minipage}
  \caption{\textbf{The necessity of dynamic state-conditioned actions and trajectory-level rewards.} (a) Action choice is state-dependent: a preliminary study on T2I-ReasonBench~\citep{sun2025t2ireasonbench} shows that different actions have respective advantages, and dynamic action making is superior to fixed workflows. (b) RL training relying only on a best-image-score reward struggles to distinguish regressive trajectories or unnecessary degradation.}
  \label{fig:teaser}
\end{figure}

Building on these observations, we propose Generation Navigator, a multi-turn T2I agent that learns to make state-conditioned actions from experience. At each turn, the navigator observes the original prompt and interaction history, then outputs a structured action consisting of an action choice $d_t\in\mathcal{A}=\{\stopact,\refineact,\regenact\}$ and a revised prompt $p_t$.
\stopact{} preserves a satisfactory image rather than risking unnecessary edits; \refineact{} edits and improves a partially correct draft in place; and \regenact{} abandons a
structurally flawed image and starts over with a revised prompt. This turns multi-turn generation from a fixed workflow into a
learnable agent problem.

To train this navigator, we introduce \pregrpo{} (Peak-Retention-Efficiency Group Relative Policy Optimization), which assigns credit to an entire multi-turn trajectory and decomposes the trajectory reward into three complementary terms: (i) Peak—discovering a high-quality image, (ii) Retention—avoiding subsequent quality degradation across turns, and (iii) Efficiency—reaching the peak in as few turns as possible.

Our contributions are summarized as follows:
\begin{itemize}
  \item We analyze the limitations of fixed T2I workflows and formulate T2I generation as a state-conditioned action-making problem.
  \item We propose Generation Navigator, a T2I agent framework, together with a trajectory data pipeline. To optimize the agent, we further introduce PRE-GRPO, a trajectory-level RL objective that jointly reduces score regression across turns and penalizes unnecessary turns.
  \item Extensive experiments on demanding benchmarks like T2I-ReasonBench and WISE demonstrate that our framework substantially outperforms existing SOTA methods.
\end{itemize}

\section{Related work}

\subsection{Text-to-image generation}
Modern T2I systems built on diffusion models, rectified flow, and autoregressive multimodal generation have improved image fidelity and prompt following \citep{rombach2022latentdiffusion,saharia2022imagen,podell2023sdxl,esser2024sd3}. Recent work further integrates language reasoning and multimodal understanding into generation pipelines \citep{chen2025januspro,deng2025bagel,qin2026unicot}. Despite these advances, single-pass generation frequently fails to faithfully satisfy challenging user intents.

\subsection{Prompt rewriting-based image generation}
Prompt-level methods leverage LLMs to optimize user prompts before generation. Early and concurrent approaches parse prompts into structured layouts, regional sub-prompts, or richer descriptions \citep{lian2024llmgrounded,yang2024rpg,kou2026thinkgenerate}, while learned rewriters such as PromptEnhancer~\citep{wang2025promptenhancer} train chain-of-thought prompt expansion with image-text alignment rewards. These methods improve intent interpretation, but they mainly operate only before generation. Our method further considers the action after generation and forms a closed loop.

\subsection{Workflow-driven closed-loop generation}
Recent systems go beyond prompt rewriting by organizing generation as multi-turn workflows. Proactive T2I agents ask clarification questions under intent uncertainty \citep{hahn2025proactiveagents}; other agent methods coordinate rewriting, self-verification, iterative improvement, and pairwise best-candidate tracking \citep{wan2025maestro,zhou2026unifiedthinker,chu2026visiondirector,chen2025t2icopilot}; dialogue-oriented systems such as DialogGen and Talk2Image address multi-turn user interaction, modality switching, intention drift, and dialogue-state consistency \citep{huang2025dialoggen,ma2025talk2image}, while Agentic Retoucher, JarvisEvo, and AgentComp study specialized repair or alignment signals \citep{shen2026agenticretoucher,lin2025jarvisevo,zarei2025agentcomp}.
However, most of these methods either adopt manually designed training-free workflows or employ fixed-workflow approaches optimized by vanilla reinforcement learning. They overlook the optimization of the T2I agent's entire trajectory. To address this issue, we propose a state-conditioned agent and utilize PRE-GRPO to enhance the trajectory-level learning capability of the T2I agent.

\section{Generation Navigator: A State-Aware T2I Agent}

As discussed in Section~\ref{sec:intro}, prompt-rewriting and fixed-workflow closed-loop systems typically yield suboptimal results because they apply the same actions regardless of the current generation state. A partially correct image may benefit from targeted refinement; a structurally flawed image is better abandoned and regenerated from scratch; and a satisfactory image should be preserved rather than exposed to unnecessary edits. The core problem is therefore not merely how to rewrite the prompt, but which structured action to output given the current state. To this end, we propose Generation Navigator, a closed-loop multi-turn T2I agent that learns to make state-conditioned actions.

\subsection{State-Conditioned Action Policy}

\begin{figure}[t]
  \centering
  \includegraphics[width=\linewidth]{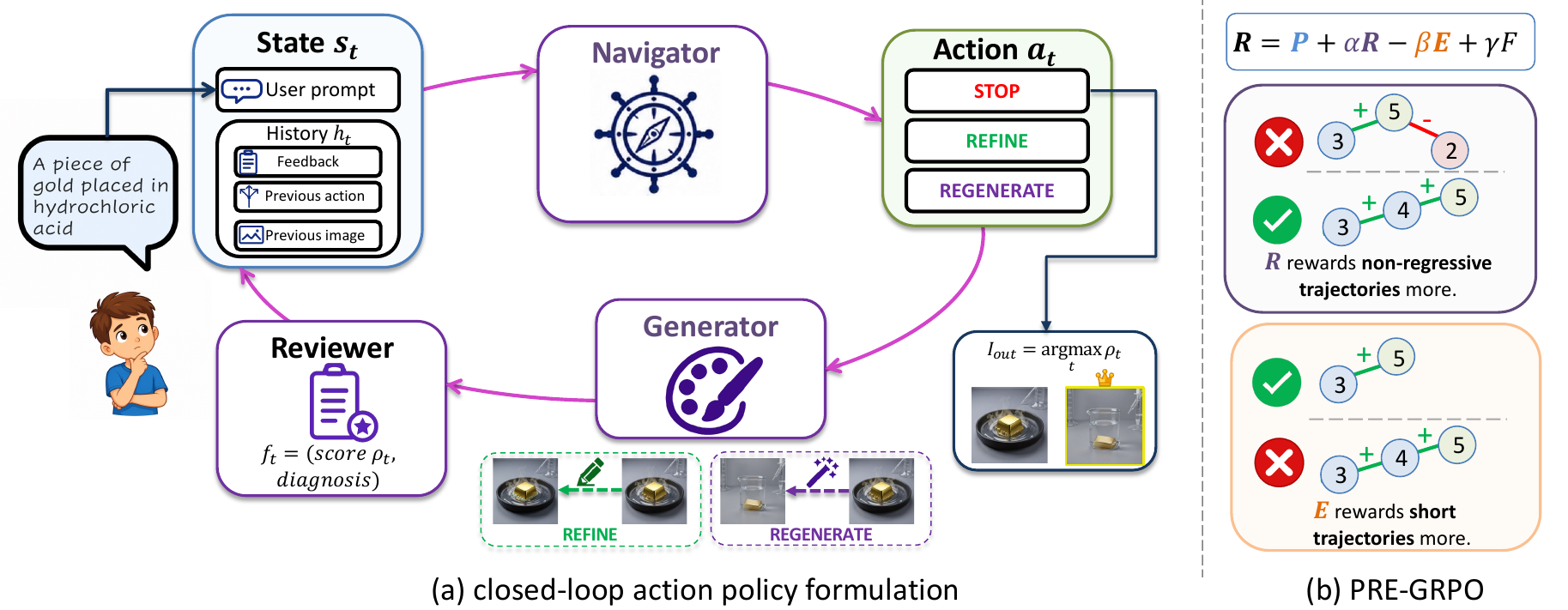}
  \vspace{-5mm}
  \caption{\textbf{Overview of Generation Navigator.} \textbf{(a)} At each turn, the navigator observes the prompt and interaction history, then outputs a structured action with one of three action choices---\stopact{}, \refineact{}, or \regenact{}---along with a revised prompt. Until the turn budget is exhausted, or \stopact{} is selected, the final output is selected as the highest-scored candidate across the entire trajectory. \textbf{(b)} PRE-GRPO decomposes the trajectory reward into (i) discovering a high-quality image ($P$), (ii) avoiding subsequent quality degradation across turns ($R$), (iii) turn efficiency ($E$), and format correctness ($F$). These terms jointly encourage efficient, non-regressive generation trajectories.}
  \label{fig:architecture}
\end{figure}

To form a closed-loop pipeline, Generation Navigator consists of three components: \textbf{navigator}, \textbf{generator}, and \textbf{reviewer}, as shown in Fig.~\ref{fig:architecture}(a). The navigator is the only learned action-making component, implemented as a multimodal LLM. The generator and reviewer serve as environment interfaces: the generator executes a generation action (text-to-image or image-to-image editing), and the reviewer evaluates the resulting image and returns textual critique together with a scalar score.

At each turn, the navigator observes the current state and outputs a structured action $a_t = (d_t,\, p_t)$, where $d_t \in \mathcal{A} = \{\stopact,\, \refineact,\, \regenact\}$ is the discrete action choice and $p_t$ is a revised prompt tailored to that choice. The initial turn and subsequent turns serve different roles. At $t{=}1$, no image or feedback is available; the navigator acts as a prompt rewriter, transforming $p_\mathrm{orig}$ into an enhanced prompt $p_1$ that is sent to the generator for text-to-image generation. From $t{\geq}2$ onward, the navigator observes the full state, consisting of the original prompt $p_\mathrm{orig}$ and the interaction history $h_t$, whose latest entry contains the generated image $I_{t-1}$ and reviewer feedback $f_{t-1}$. It outputs a state-conditioned action: \refineact{} edits the current image guided by $p_t$; \regenact{} produces a new image from scratch using $p_t$; and \stopact{} terminates exploration. It can be formulated as:

\begin{equation}
  a_t \sim \pi_\theta(\cdot \mid s_t), \quad
  s_t =
  \begin{cases}
    (p_\mathrm{orig},\; \varnothing)
      & t = 1 \\[4pt]
    (p_\mathrm{orig},\; h_t)
      & t \geq 2
  \end{cases}
\end{equation}
where $h_t = \{(a_k, I_k, f_k)\}_{k=1}^{t-1}$ is the accumulated
interaction history before turn $t$, containing all previous actions,
generated images, and reviewer feedback.
At $t{=}1$ the history is empty and the navigator sees only the original
prompt.

We record the generated candidates in a rollout as
\begin{equation}
  \tau = (s_1, a_1, I_1, f_1, \;\ldots,\; s_T, a_T, I_T, f_T),
  \quad T \leq T_\mathrm{max}.
\end{equation}
The rollout stops when the navigator selects \stopact{} at the next decision state or when the maximum turn budget $T_\mathrm{max}$ is exhausted.
Importantly, termination does not commit the system to the final
generated image; the delivered output is always the highest-scored
candidate across the entire trajectory:
\begin{equation}
  I_\mathrm{out} = I_{t^\star}, \quad
  t^\star = \arg\max_{1 \leq t \leq T}\, \rho_t.
\end{equation}
The learning objective is to find a policy that maximizes expected
trajectory return:
\begin{equation}
  \pi^\star = \arg\max_{\pi_\theta}\;
  \mathbb{E}_{\tau \sim \pi_\theta}\bigl[R(\tau)\bigr].
\end{equation}
where $R(\tau)$ is a scalar reward assigned to the entire trajectory
$\tau$. However, since the delivered image is the best candidate across
the trajectory, a reward based solely on $\rho_{t^\star}$ assigns equal
credit to every action in the rollout---including later turns that
degrade the image or waste turns after a strong candidate has already
been found. 
A good trajectory reward should therefore capture not only
whether the agent discovers a high-quality image, but also whether it
retains that quality toward the end and reaches it without unnecessary
turns. We present such a reward in Section~\ref{sec:learning}.

% \subsection{Inference-time action execution}
% At inference time, \method{} repeatedly applies the learned state-conditioned action policy until \stopact{} is selected or the turn budget is exhausted. Given state $s_t$, the navigator predicts $a_t=(d_t,p_t)$. If $d_t=\refineact$, the generator performs image-conditioned editing using the current image and the revised prompt. If $d_t=\regenact$, it restarts generation from the revised prompt. If $d_t=\stopact$, the system ends exploration and invokes the best-so-far selector defined above.

% The reviewer evaluates each newly generated image and returns textual critique together with a scalar score. These signals are not treated as oracle labels; they are observations that shape the next action. The updated state is then serialized from the original prompt, the accumulated history, the current image, and reviewer feedback. In this way, the loop is not a fixed refinement workflow. The same feedback interface can lead to different actions depending on the current visual failure mode: local defects tend to favor \refineact{}, global composition failures may favor \regenact{}, and strong candidates favor \stopact{}.

\subsection{Learning the Navigator}
\label{sec:learning}

As shown in Fig.~\ref{fig:teaser}(a), the training-free (TF) agent
already improves over fixed-workflow agents by conditioning its actions
on the current state. However, a gap remains compared to the
preference reference, indicating that actions still leave room
for improvement in both action selection and prompt rewriting.
To close this gap, we adopt a two-stage training procedure.
The first stage is supervised fine-tuning: we construct a trajectory
data pipeline yielding 103K trajectories and use them to teach
the navigator the interaction format and the semantics of each action. The second stage applies reinforcement learning to optimize
trajectory-level preferences. Since a vanilla RL reward cannot
distinguish trajectories that regress after finding a strong candidate
from those that retain quality, nor penalize unnecessary turns that
waste turns without improvement, we propose PRE-GRPO to
explicitly reward efficient, non-regressive trajectories.

\subsubsection{Action trajectory construction}
Simple prompt-image pairs are insufficient for learning an action policy because they do not contain intermediate states, structured actions, or outcomes. We therefore construct full trajectory data.
We first score the prompts in the prompt pool from multiple dimensions, including Text/symbolic reasoning, Cardinality/counting, Linguistic reasoning, Domain knowledge, Attribute binding, Counter-intuitive reasoning, and Spatial/structural reasoning. Based on the scores, we select complex samples and rewrite/enhance part of the simple samples. Afterwards, the heuristic explorer performs tree-based reasoning over the actions of \refineact{} and \regenact{}. Meanwhile, an external multimodal reviewer scores each generated image and selects the optimal branch for subsequent reasoning. Each recorded branch stores the current state, selected action, revised prompt, generated image, reviewer feedback, and subsequent scores. We further filter out regression trajectories and those with low final scores. Finally, we obtain a filtered dataset of 103K trajectories. More details can be found in the Appendix~\ref{app:data_pipeline}.

\subsubsection{Supervised fine-tuning for cold start}
After trajectory construction, we convert the filtered trajectories into multi-turn conversational format and perform supervised fine-tuning. The goal of SFT is to turn a general-purpose multimodal language model into a functional generation navigator before online policy optimization—i.e., cold-start initialization. Through standard next-token cross-entropy training, the model learns the concrete semantics of our defined action-choice space (\stopact{}, \refineact{}, \regenact{}), acquires the ability to produce the correct format, and establishes the behavioral foundation required for subsequent on-policy reinforcement learning. 

\subsubsection{\pregrpo{} for trajectory-level preference optimization}
\label{sec:pregrpo}
However, because the SFT objective maximizes the likelihood of reference actions, it is not directly optimizing for better trajectories or higher-quality generated images. The learned policy inevitably inherits biases from the constructed data, and cannot perform trajectory-level credit assignment. We therefore further optimize the navigator with reinforcement learning.

Vanilla GRPO with a best image score reward conflates unnecessary turns and regression cases as shown in Fig.~\ref{fig:teaser}(b), because it assigns the same scalar credit to all actions in a rollout regardless of what happens after the best turn. We address this with PRE-GRPO, as shown in Fig.~\ref{fig:architecture}(b) and Equation~\ref{eq:pregrpo_reward}, which decomposes trajectory quality into three complementary components.

For each training prompt $x$, the current navigator samples a group of
$K$ complete rollouts $G(x) = \{\tau_i\}_{i=1}^K$. Let $\rho_{i,t}$
denote the reviewer score at turn $t$ in rollout $\tau_i$, let $T_i$ denote
the number of generated candidates in rollout $\tau_i$, and let $\hat{\rho}_{i,t} = \rho_{i,t} / \rho_\mathrm{max}$ be the normalized
score. We compute three trajectory-level statistics:
\begin{equation}
  P_i = \max_{1 \leq t \leq T_i} \hat{\rho}_{i,t}, \quad
  R_i = \hat{\rho}_{i,T_i}, \quad
  E_i = \frac{T_i - 1}{T_\mathrm{max} - 1},
\end{equation}
where $P_i$ measures how good the best image found in the trajectory
is, $R_i$ measures whether the trajectory ends near that best image
rather than degrading afterward, and $E_i$ measures how many turns were
consumed to get there.
\begin{equation}
  R(\tau_i)=
  \underbrace{P_i}_{\text{\textbf{P}eak discovery}}
  +\alpha\cdot\underbrace{R_i}_{\text{final score \textbf{R}etention}}
  -\beta\cdot\underbrace{E_i}_{\text{turn \textbf{E}fficiency}}
  +\gamma\cdot\underbrace{F_i}_{\text{format correctness}},
  \label{eq:pregrpo_reward}
\end{equation}

where $F_i$ is the \textbf{format correctness term} that encourages well-formed actions and valid prompts. The \textbf{peak discovery term} rewards the policy for discovering a strong candidate anywhere in
the trajectory. The \textbf{retention term} discourages regression degradation: if the agent finds a good image but then continues into worse states,
$R_i$ drops well below $P_i$ and the reward decreases. The \textbf{efficiency term}
penalizes unnecessary turns, especially when the current candidate is
already satisfactory. 
In particular, the first two terms can be rewritten as:
\begin{equation}
  P_i+\alpha R_i=(1+\alpha)P_i-\alpha(P_i-R_i).
\end{equation}
The gap $P_i - R_i$ directly measures how much quality was lost between
the best image and the last generated candidate. The combined objective penalizes
this gap while preserving the incentive to find a high-quality image in
the first place.

We then normalize returns within the rollout group:
\begin{equation}
  A_i=\frac{R(\tau_i)-\mathrm{mean}_{\tau_j\in\mathcal{G}(x)}R(\tau_j)}
  {\mathrm{std}_{\tau_j\in\mathcal{G}(x)}R(\tau_j)+\epsilon}.
\end{equation}
Comparing trajectories under the same prompt removes sensitivity to
prompt difficulty and reviewer-score scale. We then apply the standard
GRPO clipped objective~\citep{shao2024deepseekmath} using $A_i$ as the
trajectory-level advantage.

Through this objective, PRE-GRPO moves credit assignment from the individual best image to
entire trajectories. Rather than rewarding or penalizing a single
generation turn in isolation, it evaluates the full trajectory the navigator
takes---whether it found a strong candidate, whether it improve
quality turn by turn, and whether it did so efficiently. This trajectory-level signal
allows the policy to learn not just what action to take at each state,
but how to plan a sequence of actions that leads to a high-quality
outcome through a short, non-regressive trajectory.
\section{Experiments}
\label{sec:experiments}

\subsection{Setup and evaluation protocol}
\label{sec:setup_evaluation_protocol}

\paragraph{Base configuration.}
Unless otherwise specified, we instantiate \method{} with Qwen3-VL-8B-Instruct as the navigator \citep{bai2025qwen3vl}, Doubao-Seed1.5 as the reviewer \citep{guo2025seed15vl}, and FLUX.2-Klein-9B as the omni-generator \citep{blackforestlabs2026flux2klein}. The same generator is used for both \refineact{} and \regenact{} actions.

\paragraph{Training data and hyperparameters.}
Our trajectory dataset contains 103K structured multi-turn trajectories constructed by the proposed data pipeline. We conducted one epoch of SFT and trained the \pregrpo{} stage for 300 optimization steps. The maximum turn budget is set to three turns. In Eq.~\ref{eq:pregrpo_reward}, we empirically set $\alpha=0.25$, $\beta=0.025$, and $\gamma=0.1$ for all main experiments. More experiments related to hyperparameters are provided in the Appendix~\ref{app:hyperparameters}.

\paragraph{Benchmarks and comparisons.}
We evaluate on three benchmarks. T2I-ReasonBench \citep{sun2025t2ireasonbench} measures compositional reasoning across four categories and reports both reasoning accuracy and image quality. WISE \citep{niu2025wise} targets knowledge-intensive generation across six domains. GenEval \citep{ghosh2023geneval} evaluates whether generated images correctly satisfy the compositional semantic requirements of prompts. Noting that these methods typically evaluate results using manually designed questions or additional detection models. This is entirely decoupled from and different from how our reviewer scores image quality.
We compare \method{} against a diverse set of baselines: proprietary one-shot systems such as GPT-4o \citep{openai2024gpt4o}, Gemini-2.0 \citep{google2024gemini20}, and SeedDream4.0 \citep{seedream2025seedream40}; open-source generators including Qwen-Image \citep{wu2025qwenimage} and FLUX.2-Klein-9B \citep{blackforestlabs2026flux2klein}; and agent baselines including GenAgent \citep{jiang2026genagent}, Think-then-Generate \citep{kou2026thinkgenerate}, IRG~\citep{huang2026interleaving}, and StruVis \citep{lyu2026struvis}.

\begin{table}[tbp]
  \caption{\textbf{Performance on T2I-ReasonBench.} \method{} is comparable to some proprietary models.}
  \label{tab:t2i_main}
  \centering
  \scriptsize
  \resizebox{\linewidth}{!}{
  \begin{tabular}{llcccccccccc}
    \toprule
    Type & Method & \multicolumn{2}{c}{Idiom} & \multicolumn{2}{c}{Textual} & \multicolumn{2}{c}{Entity} & \multicolumn{2}{c}{Scientific} & \multicolumn{2}{c}{Overall} \\
    \cmidrule(lr){3-4}\cmidrule(lr){5-6}\cmidrule(lr){7-8}\cmidrule(lr){9-10}\cmidrule(lr){11-12}
    & & Acc. & Qual. & Acc. & Qual. & Acc. & Qual. & Acc. & Qual. & Acc. & Qual. \\
    \midrule
    \graycell{Proprietary} & \graycell{Gemini-2.0~\citep{google2024gemini20}} & \graycell{52.40} & \graycell{87.80} & \graycell{73.00} & \graycell{83.30} & \graycell{67.00} & \graycell{94.30} & \graycell{66.70} & \graycell{89.30} & \graycell{64.80} & \graycell{88.70} \\
    \graycell{Proprietary} & \graycell{GPT-4o~\citep{openai2024gpt4o}} & \graycell{75.70} & \graycell{94.50} & \graycell{86.90} & \graycell{97.60} & \graycell{77.50} & \graycell{96.60} & \graycell{74.70} & \graycell{94.30} & \graycell{78.70} & \graycell{95.80} \\
    \midrule
    Open-source & SD-3.5-large~\citep{esser2024sd3} & 35.60 & 85.30 & 62.20 & 75.40 & 46.60 & 92.60 & 52.90 & 84.50 & 49.30 & 84.40 \\
    Open-source & Bagel w/CoT~\citep{deng2025bagel} & 44.60 & 84.30 & 44.00 & 73.70 & 52.40 & 91.60 & 57.70 & 88.30 & 49.70 & 84.50 \\
    Open-source & HunyuanImage-3.0~\citep{cao2026hunyuanimage30} & 25.40 & 80.20 & 54.20 & 80.90 & 52.30 & 92.20 & 56.80 & 84.40 & 47.20 & 84.40 \\
    Open-source & UniCoT~\citep{qin2026unicot} & 49.00 & 84.20 & 58.10 & 92.30 & 73.50 & 92.90 & 51.90 & 71.70 & 58.10 & 85.30 \\
    Open-source & Qwen-Image~\citep{wu2025qwenimage} & 48.10 & 84.30 & 66.50 & 85.80 & 57.10 & 84.70 & 59.50 & 85.30 & 57.80 & 87.50 \\
    Open-source & FLUX.2-Klein-9B~\citep{blackforestlabs2026flux2klein} & 55.35 & 93.83 & 43.21 & 77.02 & 57.78 & 84.29 & 74.44 & 76.12 & 57.70 & 82.82 \\
    \midrule
    Agent & Think-then-Generate~\citep{kou2026thinkgenerate} & 58.50 & 90.60 & 75.20 & 89.50 & 68.80 & 95.20 & 72.90 & 93.50 & 68.30 & 92.20 \\
    Agent & StruVis~\citep{lyu2026struvis} & 70.15 & 92.99 & 75.22 & 84.75 & 76.74 & 97.33 & 72.18 & 92.12 & 73.57 & 91.80 \\
    % Agent & Ours (w/o SFT+RL) & 67.82 & 97.79 & 64.47 & 90.56 & 73.98 & 92.62 & 85.57 & 90.17 & 72.96 & 92.79 \\
    % Agent & Ours (w/o RL) & 69.53 & 96.79 & 69.17 & 92.50 & 75.40 & 90.62 & 88.16 & 88.46 & 75.57 & 92.09 \\
    \rowcolor{blue!8}
    Agent & \method{} & 74.96 & 94.79 & 89.71 & 94.62 & 73.90 & 97.08 & 77.66 & 92.83 & 79.06 & 94.83 \\
    \bottomrule
  \end{tabular}}
\end{table}
\subsection{Main benchmark results}

Tables~\ref{tab:t2i_main} and \ref{tab:wise_main} summarize the main comparison. Across both benchmarks, our method outperforms the comparison methods.
On T2I-ReasonBench, \method{} reaches 79.06\% overall reasoning accuracy. It significantly improves over the FLUX.2-Klein-9B one-shot generator \citep{blackforestlabs2026flux2klein}, is comparable to GPT-4o \citep{openai2024gpt4o}, and is superior to recent reasoning or agentic baselines such as StruVis \citep{lyu2026struvis}. On WISE, \method{} obtains an overall score of 0.90. It exceeds proprietary models such as SeedDream4.0 \citep{seedream2025seedream40}, and outperforms closed-loop baselines such as Think-then-Generate \citep{kou2026thinkgenerate}.

\begin{table}[tbp]
  \caption{\textbf{Performance on the knowledge-intensive WISE benchmark (judged by GPT-4o).} \method{} achieves superior results across all sub-tasks compared with other baselines.}
  \label{tab:wise_main}
  \centering
  \scriptsize
  \resizebox{\linewidth}{!}{
  \begin{tabular}{llccccccc}
    \toprule
    Type & Method & Cultural & Time & Space & Biology & Physics & Chemistry & Overall \\
    \midrule
    \graycell{Proprietary} & \graycell{SeedDream4.0~\citep{seedream2025seedream40}} & \graycell{0.84} & \graycell{0.78} & \graycell{0.90} & \graycell{0.85} & \graycell{0.83} & \graycell{0.66} & \graycell{0.82} \\
    \graycell{Proprietary} & \graycell{GPT-4o~\citep{openai2024gpt4o}} & \graycell{0.81} & \graycell{0.71} & \graycell{0.89} & \graycell{0.83} & \graycell{0.79} & \graycell{0.74} & \graycell{0.80} \\
    \midrule
    Open-source & SD-3.5-large~\citep{esser2024sd3} & 0.44 & 0.50 & 0.58 & 0.44 & 0.52 & 0.31 & 0.46 \\
    Open-source & Bagel w/CoT~\citep{deng2025bagel} & 0.76 & 0.69 & 0.75 & 0.65 & 0.75 & 0.58 & 0.70 \\
    Open-source & HunyuanImage-3.0~\citep{cao2026hunyuanimage30} & 0.58 & 0.57 & 0.72 & 0.56 & 0.68 & 0.35 & 0.58 \\
    Open-source & UniCoT~\citep{qin2026unicot} & 0.76 & 0.70 & 0.76 & 0.73 & 0.81 & 0.73 & 0.75 \\
    Open-source & Qwen-Image~\citep{wu2025qwenimage} & 0.62 & 0.63 & 0.78 & 0.55 & 0.67 & 0.35 & 0.61 \\
    Open-source & FLUX.2-Klein-9B~\citep{blackforestlabs2026flux2klein} & 0.58 & 0.65 & 0.75 & 0.63 & 0.65 & 0.42 & 0.61 \\
    \midrule
    Agent & GenAgent~\citep{jiang2026genagent} & 0.78 & 0.67 & 0.78 & 0.72 & 0.71 & 0.55 & 0.72 \\
    Agent & Think-then-Generate~\citep{kou2026thinkgenerate} & 0.80 & 0.74 & 0.83 & 0.81 & 0.85 & 0.66 & 0.79 \\
    % Agent & Ours (w/o SFT+RL) & 0.83 & 0.77 & 0.92 & 0.84 & 0.81 & 0.68 & 0.82 \\
    % Agent & Ours (w/o RL) & 0.86 & 0.85 & 0.92 & 0.84 & 0.83 & \textbf{0.82} & 0.86 \\
    Agent & IRG~\citep{huang2026interleaving} & 0.78 & 0.72 & 0.76 & 0.81 & 0.82 & 0.78 & 0.77 \\
    \rowcolor{blue!8}
    Agent & \method{} & 0.93 & 0.87 & 0.96 & 0.86 & 0.87 & 0.82 & 0.90 \\
    \bottomrule
  \end{tabular}}
\end{table}

\subsection{Ablation on Action Training}
\label{sec:action_training_ablation}

\begin{figure}[tbp]
  \centering
  \includegraphics[width=\linewidth]{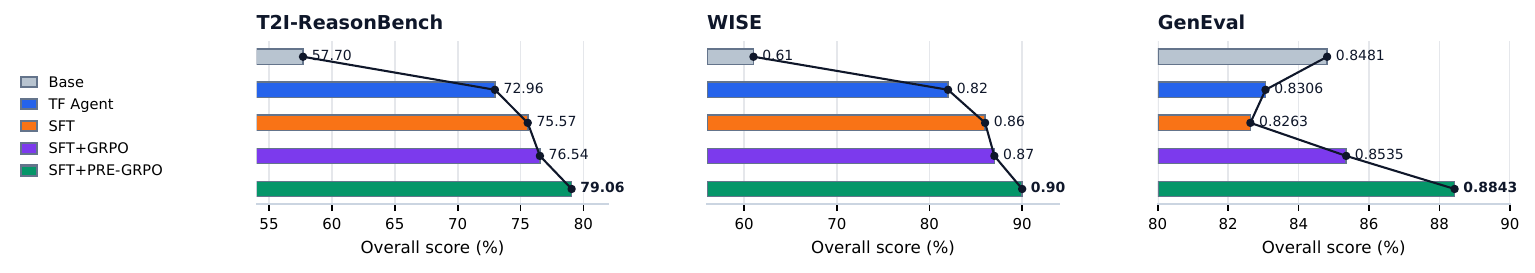}
  \caption{\textbf{Training-stage ablation reveals progressive gains on complex tasks and learned restraint on simple tasks.} For complex reasoning and knowledge-intensive tasks (T2I-ReasonBench and WISE), multi-turn action making and policy optimization yield steady improvements.  For simple tasks (GenEval), TF Agent and SFT degrade below the one-shot baseline due to the narrow action distribution caused by hand-crafted decisions and trajectory cloning. PRE-GRPO fixes this with trajectory-level rewards by pushing the policy toward suitable action choices on both simple and complex prompts and exhibits strong robustness and adaptability.}
  \label{fig:main_ablation}
\end{figure}

We first analyze the performance gains brought by each component. As shown in Fig.~\ref{fig:main_ablation}, introducing the agent framework, along with SFT and reinforcement learning, yields consistent performance improvements on both T2I-ReasonBench and WISE. Furthermore, our PRE-GRPO achieves more prominent gains compared with the vanilla GRPO algorithm.
% On the GenEval benchmark, a striking pattern emerges: Training-free (TF) Agent (0.8306) and SFT (0.8263) both fall below the one-shot baseline (0.8481). 
% This degradation exposes a fundamental limitation of heuristic multi-turn systems — over-iteration. When prompts are compositionally simple, an unregulated agent continues to refine or regenerate images that were already satisfactory, paradoxically introducing new errors. This is precisely the failure mode that motivates PRE-GRPO: its efficiency penalty explicitly discourages unnecessary turns, teaching the navigator when not to act. Quantitatively, PRE-GRPO reduces the average number of generation turns on GenEval from
% ${\sim}1.95$ (TF Agent and SFT) to 1.67, confirming that the efficiency penalty teaches
% the navigator to terminate early when the initial generation already satisfies simple compositional
% requirements (see Appendix~\ref{app:iteration_steps} for a detailed breakdown). As a result, PRE-GRPO not only recovers from the over-iteration degradation but surpasses the one-shot baseline by a substantial margin (0.8843 vs. 0.8481), while simultaneously achieving the highest scores on the complex-reasoning benchmarks T2I-ReasonBench and WISE. 
% This degradation reveals a subtler failure mode on compositionally simple tasks. 
However, on GenEval, both TF Agent and SFT fall below the one-shot baseline. 
We found that GenEval prompts are characteristically short, templated phrases such as "a photo of a cat and a dog" that specify only basic object presence or spatial relations. This input style differs from what TF Agent and SFT are adapted to: the former's hand-crafted decision rules are designed for complex scenarios, while the latter is trained on multi-turn trajectories dominated by hard prompts—both resulting in a narrow action distribution. In contrast, PRE-GRPO optimizes the entire trajectory with trajectory-level rewards, enabling the policy to adaptively calibrate its behavior to different prompts, exhibiting strong robustness and adaptability.
% GenEval prompts often involve straightforward object or relation requirements where
% few-turn generation already produces adequate candidates. For such prompts, wrong actions hurt more than they help. For example, the TF Agent may flag minor or irrelevant issues: regenerating an image that only needed minor editing, or locally refining a failure that should have been regenerated entirely. Even with best-image selection, wrong actions may still hurt the candidate pool by failing to produce anything better than the one-shot result. SFT shows a similar pattern: it imitates all demonstrated actions equally without reward signal, so it can reproduce suboptimal choices on easy prompts.
% PRE-GRPO fixes this with trajectory-level rewards: the retention and efficiency terms penalize actions that do not improve the current state, pushing the policy toward better action choices on both simple and complex prompts while avoiding low-value actions. The average number of turns in Appendix~\ref{app:iteration_steps} further corroborates this point indirectly.

\subsection{PRE-GRPO analysis}

\begin{figure}[tbp]
  \centering
  \includegraphics[width=\linewidth]{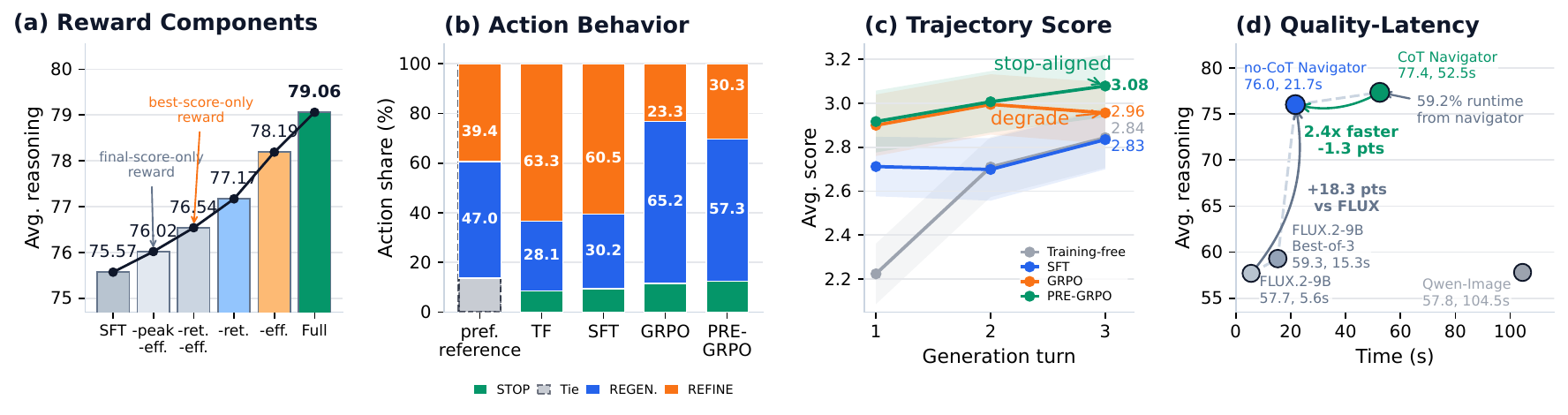}
  \caption{\textbf{\pregrpo{} action analysis and quality--latency trade-off.} (a) \textbf{Reward components:} \pregrpo{} outperforms final-score-only and best-score-only reward variants, and both the retention and efficiency terms contribute to the final performance. Here, ``-peak'', ``-ret.'' and ``-eff.'' denote reward variants w/o the peak, retention, and efficiency terms, respectively. (b) \textbf{Action behavior:} \pregrpo{} shifts the action distribution closer to the empirical preference reference, balancing decisive regenerations and targeted refinements. (c) \textbf{Trajectory score:} the score of PRE-GRPO increases monotonically at each turn, yielding superior trajectories. (d) \textbf{Quality--latency trade-off:} Our method can achieve a quality-latency balance by removing CoT.}
  \label{fig:pregrpo_path_analysis}
\end{figure}

\paragraph{Reward ablation.}
As shown in Fig.~\ref{fig:pregrpo_path_analysis}(a), PRE-GRPO achieves substantial performance gains compared with GRPO with a final-score-only reward and GRPO with a best-score-only reward. Meanwhile, removing either the retention term or the efficiency term in Equation~\ref{eq:pregrpo_reward} leads to performance degradation. This verifies that the two terms can effectively facilitate action-policy learning.

\paragraph{Action behavior.}
Fig.~\ref{fig:pregrpo_path_analysis}(b) shows how each training stage reshapes
the action distribution. TF Agent and SFT both deviate
significantly compared to the preference reference, heavily favoring edits (\refineact{}
accounts for over 60\% of actions), yet this bias does not translate into better
performance. Vanilla GRPO overcorrects in the opposite direction by using 65.15\%
\regenact{}, becoming overly aggressive about restarting. PRE-GRPO achieves a
more balanced distribution that is closest to the preference reference.

\paragraph{Trajectory score.}
Fig.~\ref{fig:pregrpo_path_analysis}(c) evaluates the average score of each turn. Vanilla GRPO finds a strong intermediate candidate but then regresses to a worse final state, because best-score-only rewards credit all actions equally and cannot distinguish those that improve quality from those that later degrade it. In contrast, PRE-GRPO maintains a steadily improving trajectory. 

\subsection{Robustness and transfer}
\label{sec:robustness_transfer}

\begin{figure}[tbp]
  \centering
  \includegraphics[width=\linewidth]{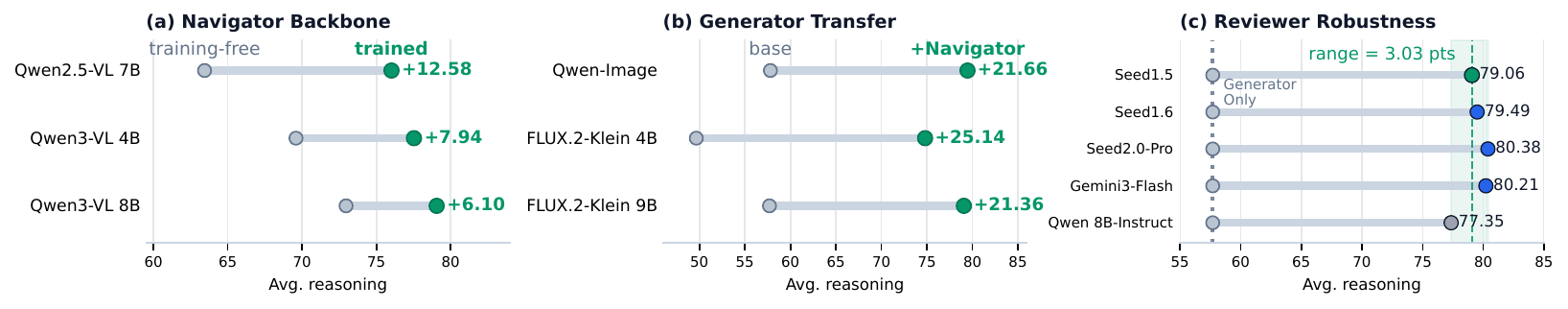}
  \caption{\textbf{Robustness and transfer across system components.} The learned state-conditioned action policy shows consistent transfer across navigator, generator, and reviewer choices. (a) \textbf{Navigator backbone:} training improves state-conditioned action-making performance across 4B, 7B and 8B navigators, indicating the robustness of our training method across different navigators. (b) \textbf{Generator transfer:} The navigator trained on Flux.2-9B can transfer to different underlying generators, which demonstrates that our method does not overfit to a single generator model. (c) \textbf{Reviewer robustness:} when we replace the reviewer used during training with a different one in inference, the performance only fluctuates within a range of 3 points, preserving a large gain over one-shot generation. This verifies that our method is not affected by the bias of a single reviewer.}
  \label{fig:component_analysis}
\end{figure}

We further test whether the learned state-conditioned action policy depends on a particular navigator backbone, generator, or reviewer. Fig.~\ref{fig:component_analysis} summarizes these component-level variations.

\paragraph{Navigator backbone.}
Fig.~\ref{fig:component_analysis}(a) shows that trajectory-level training improves state-conditioned action-making capability across navigator backbones and scales. The gains are not limited to the largest navigator: Qwen3-VL-4B improves from 69.59\% to 77.53\%, while Qwen3-VL-8B improves from 72.96\% to 79.06\% \citep{bai2025qwen3vl}. This indicates that our training paradigm is effective across different MLLMs, demonstrating the generalization ability of our method.

\paragraph{Generator transfer.}
Fig.~\ref{fig:component_analysis}(b) evaluates whether the learned navigator transfers when the underlying image generator changes. Specifically ,we directly transfer the navigator trained on FLUX.2-9B to different generators for evaluation. The same navigator can improve both Qwen-Image \citep{wu2025qwenimage} and FLUX-family generators \citep{blackforestlabs2026flux2klein}, with the largest relative benefit on the weaker FLUX.2-Klein-4B generator (49.66\% to 74.80\%). This demonstrates that our method effectively learns how to perform action-making, rather than overfitting to the pattern of a single generator.

\paragraph{Reviewer robustness and bias analysis.}
We further investigate whether the navigator's competence is merely a proxy for the reviewer's capability. To test this, we fix the navigator (trained exclusively with Seed1.5 feedback) and swap the inference-time reviewer across four alternatives spanning different architectures and capability tiers: Seed1.6, Seed2.0-Pro, Gemini3-Flash, and Qwen3-VL-8B-Instruct~\cite{bai2025qwen3vl}. \textbf{In this case, the reward model for training, the reviewer for inference, and the benchmark evaluator are completely distinct models.} As Fig.~\ref{fig:component_analysis}(c) shows, performance remains within a narrow band of merely 3.03 points (77.35\%–80.38\%). Notably, even replacing the proprietary Seed1.5 with an entirely different architecture (Gemini3-Flash) yields 80.21\% — actually exceeding the training-time reviewer's result. This confirms that our navigator is highly robust and generalizable, capable of making effective action choices based on diverse feedback and current state. This action-making skill is therefore architecture-agnostic: the navigator learns how to act, not what a specific reviewer prefers.

\subsection{Efficiency}
\label{sec:efficiency_human_validation}
Agentic generation introduces additional latency, so Fig.~\ref{fig:pregrpo_path_analysis}(d) reports the quality-latency trade-off on T2I-ReasonBench by using Qwen3-VL-8B-Instruct as a reviewer for runtime measurement (results of WISE are in Appendix~\ref{app:wise_latency}). The full-CoT
Navigator reaches 77.4\% accuracy at 52.5s per example. However, we found that 60\% of the overhead stems from the Navigator because of its long Chain-of-Thought (CoT). For a faster alternative, we train a no-CoT Navigator that achieves 76.0\% accuracy with an inference latency of 21.7 seconds per sample. It is 2.4x faster than the full-CoT Navigator while still substantially outperforming both the FLUX.2-Klein-9B generator-only baseline and its Best-of-3 variant. This demonstrates a practical quality-latency balance. Appendix~\ref{app:latency_tradeoff} further shows that our method gains over the baseline even with fewer max turns, allowing users to flexibly balance quality and latency.

\section{Conclusion}
We presented Generation Navigator, an agentic framework motivated by the
observation that the optimal next action in multi-turn T2I generation is inherently
state-dependent—yet existing methods apply fixed workflows regardless of the
current image quality. Our framework recasts this process as a learnable,
state-conditioned action-making problem, replacing hand-designed heuristics with a
data-driven policy. To address trajectory-level credit assignment, we introduced
PRE-GRPO, which jointly rewards peak quality, score retention, and turn efficiency.
Experiments across multiple benchmarks confirm substantial gains over existing
one-shot and agentic baselines.

\newpage
\bibliographystyle{plainnat}
\bibliography{main}

\newpage
\appendix

\section{Appendix Roadmap}

The appendix provides reproducibility details, metric definitions, qualitative evidence, and a human validation study that complement the main paper:
\begin{enumerate}
  \item \textbf{Pilot study details.} Appendix~\ref{app:pilot_study} expands the setup behind the introductory pilot study, including the three-turn workflow and the fixed-action versus preference-reference comparison.
  \item \textbf{Data pipeline and distribution.} Appendix~\ref{app:data_pipeline} describes the construction of the 103K structured trajectories, including prompt-pool scoring, targeted prompt augmentation, branch-and-select exploration, and trajectory filtering. This section supports the training-data claims in Section~\ref{sec:setup_evaluation_protocol} and clarifies how action-trajectory supervision differs from ordinary prompt-image data.
  \item \textbf{Agent prompts.} Appendix~\ref{app:agent_prompts} gives the complete reviewer and navigator prompt templates.
  \item \textbf{Hyperparameter studies.} Appendix~\ref{app:hyperparameters} analyzes the reward weights $\alpha$ and $\beta$ in Eq.~\ref{eq:pregrpo_reward}, the turn budget $T_{\max}$, and representative reward-scale examples. These studies explain why the default \pregrpo{} configuration balances candidate discovery, terminal retention, and turn efficiency.
  \item \textbf{Controlled sampling-budget comparison.} Appendix~\ref{app:sampling_budget_comparison} compares one-shot generation, best-of-3 selection, prompt enhancement, fixed-workflow agents, and trained action-policy variants under a single T2I-ReasonBench view. This isolates the net contribution of state-conditioned action making from gains caused by extra sampling budget.
  \item \textbf{Best-score versus final-score selection.} Appendix~\ref{app:best_vs_final_selection} ablates the interaction between trajectory-level reward choice and inference-time output selection, explaining why \method{} returns the highest-scored candidate along the trajectory.
  \item \textbf{Average generation turns.} Appendix~\ref{app:iteration_steps} reports the average number of generation turns on GenEval as an indirect signal of action calibration on simple compositional prompts.
  \item \textbf{Quality--latency trade-off.} Appendix~\ref{app:latency_tradeoff} studies how the maximum turn budget affects accuracy and latency for the no-CoT Navigator.
  \item \textbf{WISE latency analysis.} Appendix~\ref{app:wise_latency} compares the quality--latency trade-off on WISE against generator-only and agent baselines.
  % \item \textbf{Stop-eligible logic and correct-stop rate.} Appendix~\ref{app:correct_stop} formalizes the trajectory-level stopping metric. It specifies the score threshold for stop eligibility, how premature stopping, delayed stopping, and budget termination are counted, and reports the correct-stop rate for each state-conditioned action policy.
  \item \textbf{Qualitative visualizations.} Appendix~\ref{app:qualitative_visualizations} presents representative multi-turn cases across textual image design, counting, spatial relations, scientific reasoning, and two-object generation. The examples expose the actual prompts, reviewer feedback, actions, and selected images behind the aggregate results.
  \item \textbf{Human evaluation.} Appendix~\ref{app:human_evaluation} reports a pairwise human study comparing human preferences with reviewer-induced preferences, providing an empirical check that the automatic reviewer is a useful signal for \pregrpo{} trajectory optimization.
  \item \textbf{Limitations and future directions.} Appendix~\ref{app:limitations_future} discusses computational trade-offs, and the use of reviewers as environment signals.
\end{enumerate}

\section{Pilot Study Details}
\label{app:pilot_study}

\paragraph{Pilot study details (Section~\ref{sec:intro}).}
We construct a three-turn workflow on T2I-ReasonBench to compare action strategies. In the first turn, each prompt is rewritten by Doubao-Seed1.5 and then fed to FLUX.2-Klein-9B for initial image generation. In subsequent turns, three action strategies diverge: (i)~\emph{Refine-only} applies image-to-image editing on the current image at every turn; (ii)~\emph{Regenerate-only} uses the MLLM (Doubao-Seed1.5) to rewrite the prompt and performs text-to-image generation from scratch at every turn; (iii)~\emph{Preference reference} executes both actions in parallel at each turn, scores both outcomes with the benchmark evaluator, and proceeds along the higher-scoring branch into the next turn.

The preference reference thus provides an empirical upper bound on greedy per-turn action selection without any learned policy. Results in Fig.~\ref{fig:teaser}(a) show that neither Refine nor Regenerate dominates---Regenerate wins 47.01\% of head-to-head comparisons, Refine wins 39.38\%, and 13.61\% are ties---confirming that the optimal action is state-dependent and motivating a learned action policy.

\section{Data Pipeline and Distribution}
\label{app:data_pipeline}

This section details how we construct the 103K structured multi-turn trajectories used for navigator training. We construct this data to support the subsequent cold-start training of SFT.

\subsection{Prompt Pool and Complexity Scoring}

\begin{figure}[h]
  \centering
  \includegraphics[width=\linewidth]{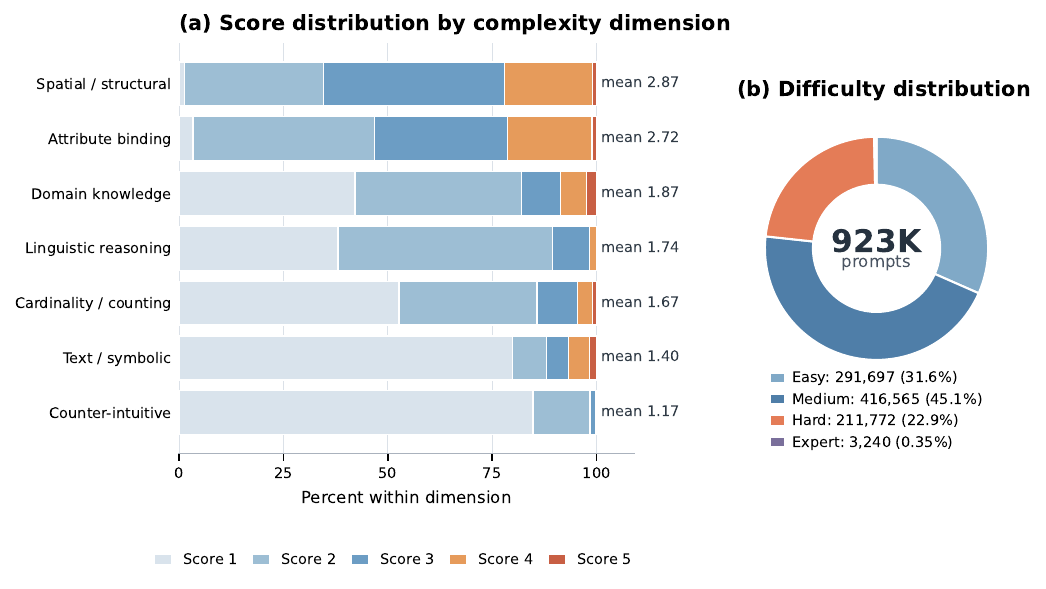}
  \caption{Prompt-pool distribution used for data construction. (a) Per-dimension score distribution over the 923K scored prompt pool, with mean score annotated on the right. (b) Overall difficulty distribution.}
  \label{fig:app_data_distribution}
\end{figure}

The first stage starts from 923,274 candidate prompts. Each prompt is scored along seven complexity dimensions: spatial and structural composition, attribute binding, cardinality and counting, counter-intuitiveness, text and symbolic content, domain knowledge, and linguistic reasoning. We use these scores to identify prompts that are likely to require nontrivial state-conditioned actions during generation. As shown in Fig.~\ref{fig:app_data_distribution}, the raw pool is skewed toward Easy/Medium/Hard prompts, while Expert cases and score-5 tails are rare. In particular, we sample complex seeds from Hard/Expert prompts and from prompts whose score is at least 4 in any complexity dimension. We also sample Easy/Medium prompts as rewriting seeds for the augmentation stage, so that the final data contain both intrinsically difficult prompts and simpler prompts transformed into different scene cases. This endows our samples with diversity.

\subsection{Prompt Augmentation}

In the second stage, we employ an LLM to rewrite Easy/Medium samples to enrich domains with insufficient hard samples. Meanwhile, we augment idiom and scientific samples that are difficult to obtain in daily scenarios, ensuring broader coverage of our dataset. The rewriting prompt is presented as follows:

\paragraph{Idiom prompt generation rules.}
The rule template used for idiom prompt generation is shown below.
\begin{lstlisting}[style=promptbox]
Track-specific rules for idiom_interpretation:
- Produce one compact everyday sentence that uses an idiom or metaphor in context.
- Usually 8-18 words, with at most one short setup clause before the idiom.
- Do not explain the figurative meaning.
- Keep it concrete enough that a later model can visualize it.
- The sentence may borrow a motif from the seed, but it should read like ordinary language.
\end{lstlisting}

\paragraph{Scientific prompt generation rules.}
The rule template used for scientific prompt generation is shown below.
\begin{lstlisting}[style=promptbox]
Track-specific rules for scientific_reasoning:
- Convert the seed into a short scientific prompt.
- Usually keep prompts within 2-22 words.
- Do not describe a generic short scene.
- Keep explanations implicit. Do not explain why something happens.
- Before finalizing each scientific prompt, silently check: is this a setup or comparison, not a revealed result or decorative scene?
- If the seed is scenic, surreal, or decorative, discard that surface framing and keep only one motif that can become a benchmark-like scientific prompt.
\end{lstlisting}

\subsection{Branch-and-select Trajectory Construction}

The third stage constructs multi-turn action trajectories with a branch-and-select procedure. Algorithm~\ref{alg:trajectory_construction} summarizes the implementation used to construct trajectory logs. At each turn, the explorer expands multiple candidate actions, scores their generated images, and continues from the best branch.

\begin{algorithm}[h]
\caption{Branch-and-select trajectory construction}
\label{alg:trajectory_construction}
\begin{algorithmic}[1]
\State \textbf{Input:} request $x$, navigator $N$, generator $G$, reviewer $R$, branch size $K$, budget $T_{\max}$, threshold $\rho_{\mathrm{thr}}$
\State $\mathcal{P}_1\gets\emptyset$, $\mathcal{H}_1 \gets \emptyset$, $\rho_{\mathrm{best}}\gets -\infty$
\For{$t=1$ \textbf{to} $T_{\max}$}
  \State $\mathcal{A}_t \gets N(x,\mathcal{P}_t)$, where $\mathcal{A}_t=\{a_t^1,\ldots,a_t^K\}$
  \For{$k=1$ \textbf{to} $K$}
    \State $I_t^k \gets G(a_t^k)$
    \State $(s_{\mathrm{visual}}^k,s_{\mathrm{instruction}}^k,c_t^k) \gets R(x,I_t^k)$
    \State $\rho_t^k \gets 0.3\,s_{\mathrm{visual}}^k+0.7\,s_{\mathrm{instruction}}^k$
  \EndFor
  \State $k^\star \gets \arg\max_{1\leq k\leq K}\rho_t^k$
  \State $\mathcal{H}_{t+1}\gets\mathcal{H}_t\cup\{(a_t^k,I_t^k,\rho_t^k,c_t^k)\}_{k=1}^{K}$
  \If{$\rho_t^{k^\star}\geq\rho_{\mathrm{thr}}$ \textbf{or} $t=T_{\max}$ \textbf{or} $\rho_t^{k^\star}\leq\rho_{\mathrm{best}}$}
    \State \Return $\mathcal{H}_{t+1}$
  \EndIf
  \State $\rho_{\mathrm{best}}\gets \max(\rho_{\mathrm{best}},\rho_t^{k^\star})$
  \State $\mathcal{P}_{t+1}\gets\mathcal{P}_t\cup\{(a_t^{k^\star},I_t^{k^\star},\rho_t^{k^\star},c_t^{k^\star})\}$
\EndFor
\end{algorithmic}
\end{algorithm}

The selected branch defines the next state for continued exploration. Note that while $\mathcal{H}_{t+1}$ logs the full tree of candidate actions and critiques for dataset construction, the navigator $N$ is conditioned only on the active selected trajectory $\mathcal{P}_t$ during rollout. In our implementation, the stopping threshold is empirically set to $\rho_{\mathrm{thr}}=4.5$ on the 0--5 reviewer-score scale. 

\subsection{Trajectory Filtering}

The heuristic explorer is used to generate broad action coverage, not as an oracle policy to be cloned directly. Before converting rollouts into SFT trajectories, we apply two filtering criteria: (1) we retain only trajectories whose best reviewer score exceeds 4.5 on the 0–5 scale, and (2) within each trajectory, we keep only the branches that show strict monotonic score improvement across turns. Branches that degrade or plateau are discarded. After filtering, the final dataset contains 103K structured multi-turn trajectories with state $s_t$, feedback $f_t=(c_t,\rho_t)$, target action $a_t=(d_t,p_t)$, and trajectory-level scores.

Note that the SFT model does not imitate the heuristic explorer wholesale; it only sees the successful trajectories that survived filtering. \pregrpo{} then further refines action preference at the trajectory level.

\subsection{Training Data Contamination Analysis}
\label{app:contamination}

Since our training trajectories are collected from an internal prompt pool, we verify that no benchmark prompt has leaked into the training data. We adopt a multi-granularity protocol combining semantic and lexical metrics:

\begin{enumerate}[leftmargin=*]
    \item \textbf{Embedding cosine similarity.} We encode each benchmark prompt and each training prompt with the \texttt{sentence-transformers/all-MiniLM-L6-v2} encoder \citep{reimers2019sentencebert,wang2020minilm} and report the maximum cosine similarity between each benchmark prompt and its nearest neighbor in the training pool. This captures paraphrase-level semantic overlap.
    \item \textbf{5-gram Jaccard index.} Measures symmetric lexical overlap at the 5-gram level.
    \item \textbf{5-gram containment.} Measures the fraction of a benchmark prompt's 5-grams that appear in its nearest training prompt, capturing asymmetric substring inclusion.
    \item \textbf{8-gram containment.} Following the contamination protocol used in PaLM \citep{chowdhery2023palm}, we flag a benchmark sample as potentially contaminated if $\geq$70\% of its 8-grams are contained in any training prompt.
    \item \textbf{13-gram collision.} Following GPT-3 \citep{brown2020language}, we check for exact 13-gram matches between benchmark and training prompts.
\end{enumerate}

\begin{figure}[h]
\centering
\includegraphics[width=\linewidth]{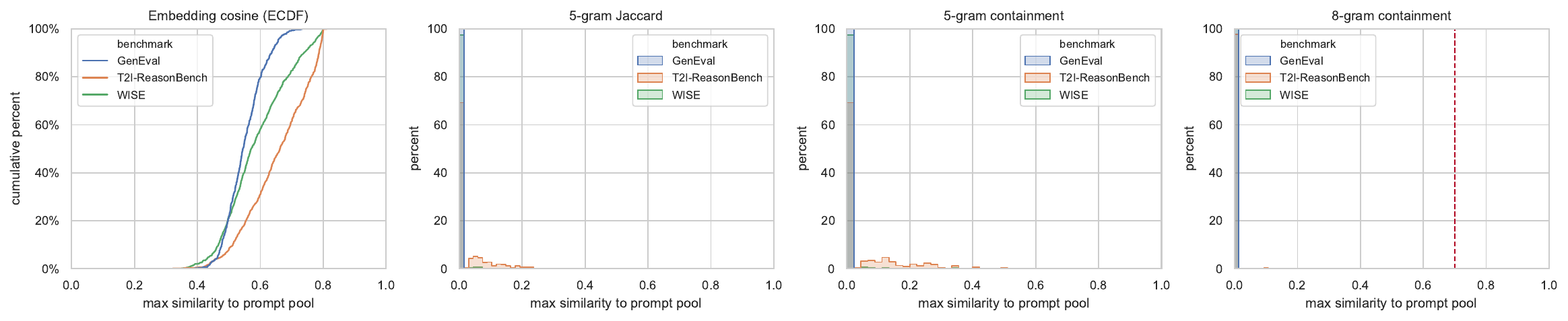}
\caption{Training data contamination analysis. From left to right: embedding cosine similarity (ECDF), 5-gram Jaccard (histogram), 5-gram containment (histogram), and 8-gram containment (histogram) with the 70\% threshold from the PaLM contamination protocol \citep{chowdhery2023palm} shown as a dashed line. No benchmark prompt exceeds the flagging threshold under any lexical metric.}
\label{fig:contamination}
\end{figure}

Fig.~\ref{fig:contamination} reports the results for the first four metrics across all three evaluation benchmarks (GenEval, T2I-ReasonBench, WISE).
The leftmost panel plots the embedding cosine ECDF. Nearest-neighbor similarities stay below the 0.8 semantic near-duplicate threshold. T2I-ReasonBench shows relatively higher values because its prompts frequently describe common natural phenomena in generic terms; the lexical metrics confirm these overlaps are not surface-level copies. Under both 5-gram Jaccard and 5-gram containment, nearly all benchmark prompts have near-zero similarity to the training pool. In the 8-gram containment panel, no prompt approaches the 70\% threshold of the PaLM contamination protocol \citep{chowdhery2023palm}. Under the 13-gram collision protocol of GPT-3 \citep{brown2020language}, we find zero exact matches.

We conclude that our results are not inflated by benchmark contamination.

\section{Agent Prompts}
\label{app:agent_prompts}

This section provides the complete prompt templates used by the reviewer and navigator. We use longer, more instructional prompts during data construction to make heuristic rollouts stable and interpretable. After trajectory data are built, the trained navigator is driven by a shorter prompt during SFT, RL, and inference, reducing token overhead while preserving the same action semantics.

\paragraph{Reviewer prompt.}
The reviewer prompt used during data construction and inference is shown below. It returns a textual diagnosis and two scalar scores. The scalar reviewer score used for ranking candidates is $\rho=0.3\cdot\text{Visual Quality}+0.7\cdot\text{Instruction Comprehension}$.
\begin{lstlisting}[style=promptbox]
**Role**: You are a strict **Image Quality Critic and QA Specialist**.

**Task**:
1. Analyze the **Current Image** strictly against the **User Request**. Keep the thinking process concise (within 512 tokens).
2. Provide a detailed diagnosis of flaws.

**Inputs**:
- **User Request**: {user_request}
- **Current Image**: (Visual Input)

**Evaluation Criteria**:
**1. Aesthetic Quality**:
**Aesthetic & Technical Quality Scoring Rules (0.0-5.0)**:
Evaluate the overall aesthetic appeal of the image and provide a score:

Assess the image for technical flaws (artifacts, anatomy distortions, blurriness) and aesthetic appeal (lighting, composition).
- **5 (Excellent):** Flawless. Professional lighting, perfect anatomy (hands/faces), sharp details, high aesthetic value.
- **4 (Good):** High quality, but may have negligible artifacts (e.g., slightly odd background texture) or average lighting.
- **3 (Acceptable):** Noticeable flaws (e.g., slightly distorted hands, weird physics, over-saturation) but the image is coherent.
- **2 (Poor):** Obvious artifacts, severe anatomy issues (extra fingers, melted faces), low resolution, or blurry.
- **1 (Unusable):** Complete noise, broken geometry, or unrecognizable visual garbage.

**2. Fulfillment (Crucial)**:
**Completeness Fulfillment (Coverage) Scoring Rules (0.0-5.0)**:
Evaluate the completeness of the image based on the requirements and provide a score:

How accurately does the image reflect the elements, attributes, style, and spatial relationships described in the prompt?
- **5 (Perfect):** All elements, modifiers, and styles are present and accurate. No missing objects.
- **4 (High):** Key elements are correct. Very minor details (e.g., a small background object or exact shade of color) might be slightly off, but the main intent is perfectly captured.
- **3 (Moderate):** Main subject is present, but some modifiers (e.g., "holding a red cup" becomes "holding a blue cup") or styles are incorrect.
- **2 (Low):** Significant deviations. Key subjects missing or incorrectly interacting. The image vaguely resembles the prompt but fails on specifics.
- **1 (Failure):** Completely irrelevant to the prompt or hallucinates entirely different subjects.

**Output Format**:
Provide a valid JSON object.
```json
{
    "evaluation": {
        "Instruction_Comprehension": [0-5 float],
        "Visual_Quality": [0-5 float]
    },
    "diagnosis": "Short objective description of flaws. Be specific. E.g., 'The lighting is flat', 'The cat is missing the hat', 'The style is 3D instead of anime'."
}
```
\end{lstlisting}

\paragraph{Navigator prompt for data construction.}
During data construction, the navigator prompt is intentionally verbose so that the heuristic agent produces diverse but interpretable rollouts.
\begin{lstlisting}[style=promptbox]
**Role**: You are an expert **Art Director and Prompt Strategist**.

Keep the thinking process concise (within 384 words).

**Context**: A Critic has evaluated the previous generation and found issues. You have final authority to choose the next action.

**Inputs**:
- **Original User Request**: {user_request}
- **Current Prompt Used**: {current_prompt}
- **Critic's Visual Quality Score**: {visual_score}/5
- **Critic's Instruction Comprehension Score**: {instruction_score}/5
- **Critic's Diagnosis**: "{diagnosis}"
- **Visual Context**: (See attached image)

**Task 1: CHOOSE the Action**
Analyze the Diagnosis and Image to choose ONE action:
- **STOP**: The image is excellent and meets the User Request perfectly.
- **REGENERATE (T2I)**: Major failures. Examples: Wrong subject, garbled composition, wrong art style, severe anatomy issues. Needs a fresh start.
- **REFINE (I2I)**: The foundation is solid, but specific details need editing. Examples: Wrong clothing color, missing specific item, bad hands, unwanted object.

**Task 2: EXECUTE the Prompt Writing**

### Scenario A: If REGENERATE (T2I)
**Strategy Shift**: The previous prompt FAILED. Do NOT just copy it or make minor tweaks.
1. **Rephrase**: Describe the subject using different adjectives or synonyms.
2. **Reorder**: Move the missing or distorted elements mentioned in the `diagnosis` to the **very beginning** of the prompt.
3. **Simplify vs. Enrich**:
   - If diagnosis says "garbled/messy" -> **Simplify** details, focus on main subject.
   - If diagnosis says "boring/wrong style" -> **Enrich** with style modifiers (e.g., "cinematic lighting", "concept art").

### Scenario B: If REFINE (I2I)
Write a specific **Editing Instruction**.
1. **Format**: Use "Add [obj]", "Remove [obj]", "Change [obj] to [obj]", or "Make [obj] [action]".
2. **Specificity**: Explicitly state the target. AVOID "it/him/her". Use "the panda", "the red car".
3. **Single Focus**: Target **ONLY ONE** specific area mentioned in the `diagnosis`. Do not fix everything at once.
4. **Avoid Anatomy**: Do NOT try to fix eyes/gaze via I2I. Use REGENERATE for that.
5. **Example**: "Make the tall warrior wear a red cape." (NOT "Change clothes").

**Task 3: Output**
Only provide a valid JSON object.
```json
{
    "decision": "STOP" | "REGENERATE" | "REFINE",
    "reasoning": "Explain why you chose this action based on the diagnosis.",
    "revised_prompt": "String OR null. If STOP, null. If REGENERATE, the full new T2I prompt. If REFINE, the specific I2I instruction."
}
```
\end{lstlisting}

\paragraph{Navigator prompt for training and inference.}
After collecting trajectory data, we replace the long construction prompt with the following compact prompt:
\begin{lstlisting}[style=promptbox]
You are an Art Director and Prompt Strategist. Based on diagnosis and current result, respond with strict JSON using keys decision and revised_prompt. The decision field must be STOP, REGENERATE, or REFINE.
\end{lstlisting}
The data-construction prompt and inference prompt differ in length and explicit guidance, but they share the same action space and output contract: stop, regenerate with a full revised T2I prompt, or refine with a localized I2I instruction.

\section{Hyperparameter Studies}
\label{app:hyperparameters}

We study three practical hyperparameters that affect action-policy learning: the terminal-retention weight $\alpha$, the turn-cost weight $\beta$ in Eq.~\ref{eq:pregrpo_reward}, and the maximum turn budget $T_{\max}$. The format correctness weight $\gamma$ is kept fixed because it mainly guards structured output validity rather than the trajectory-quality trade-off. To reduce sweep cost, these ablations use 100-step \pregrpo{} runs while varying each $\alpha$ or $\beta$ factor at a time; the main model reported in Section~\ref{sec:experiments} is trained for 300 steps.

\begin{figure}[tbp]
  \centering
  \includegraphics[width=\linewidth]{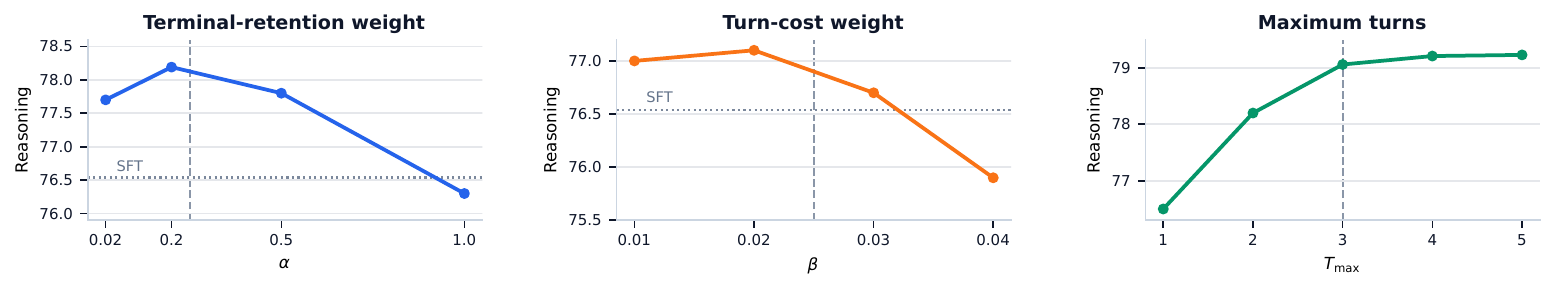}
  \caption{\textbf{Hyperparameter studies.} We sweep the terminal retention weight $\alpha$, turn-cost weight $\beta$, and maximum turn budget $T_{\max}$. Dashed vertical lines mark the default setting used in the main experiments. The dotted horizontal baseline in the first two panels denotes the corresponding SFT baseline. Both $\alpha$ and $\beta$ consistently outperform the baseline over a wide range of values, demonstrating that our method is robust to hyperparameters. Within this broad interval, both $\alpha$ and $\beta$ can effectively impose path constraints.}
  \label{fig:app_hyperparameter_sweeps}
\end{figure}

\paragraph{Reward weights.}
Fig.~\ref{fig:app_hyperparameter_sweeps} shows that $\alpha$ remains strong from 0.02 to 0.5, while larger values such as 1.0 reduce performance. In the useful range, the score-retention term acts as a stabilizer for stop alignment without disturbing peak discovery. When $\alpha$ becomes too large, the policy can overemphasize score retention and lose some exploratory benefit. The turn-cost weight $\beta$ behaves as a weak efficiency regularizer: a small penalty is sufficient to prefer shorter trajectories when quality is comparable, while an overly large penalty suppresses useful exploration.

\paragraph{Scale intuition.}
The default weights are intended to regularize action selection rather than to override candidate discovery. For readability, let $y=[\rho_1,\ldots,\rho_T]$ be a raw reviewer-score sequence and ignore the fixed format correctness term. With the default $T_{\max}=3$, $\alpha=0.25$, and $\beta=0.025$, Eq.~\ref{eq:pregrpo_reward} reduces to the following normalized comparison:
\[
  \widetilde R(y)
  = \frac{\max_t \rho_t}{5}
  +0.25\cdot\frac{\rho_T}{5}
  -0.025\cdot\frac{T-1}{2}.
\]
Each extra turn costs $0.025/(T_{\max}-1)=0.0125$ in normalized reward under $T_{\max}=3$; using all three turns instead of one therefore costs 0.025, equivalent to roughly 0.10 raw reviewer-score points when peak and terminal scores move together. This scale yields the following representative preferences, where numbers in parentheses are $\widetilde R$ values:
\[
\begin{aligned}
  \text{(i)}\quad [4.80]\;(1.2000) &> [4.50,4.80]\;(1.1875),\\
  \text{(ii)}\quad [3.00,4.00,4.80]\;(1.1750) &> [3.00,4.80,4.00]\;(1.1350).
\end{aligned}
\]
These comparisons show that the reward prefers a shorter trajectory at equal quality and, for trajectories with the same peak and length, prefers ending near the discovered peak.
Thus, $\alpha$ and $\beta$ regularize the state-conditioned action policy toward stable and economical trajectories without turning the objective into terminal-only or length-only optimization.

The same calculation explains why larger weights hurt. If $\beta$ is increased to 0.05, the length penalty can become too aggressive:
\[
  \widetilde R_{\beta=0.05}([4.83])=1.2075
  \;>\;
  \widetilde R_{\beta=0.05}([4.0,4.6,5.0])=1.2000,
\]
so the reward would reject a trajectory that eventually reaches a substantially better candidate. Similarly, if $\alpha$ is made much larger, terminal retention can make lower-peak stable trajectories overtake trajectories that discover a stronger peak:
\[
  \widetilde R_{\alpha=4.0}([3.00,4.50,4.70])=4.6750
  \;>\;
  \widetilde R_{\alpha=4.0}([3.00,4.90,4.64])=4.6670.
\]
This matches the sweep trend in Fig.~\ref{fig:app_hyperparameter_sweeps}: overly large terminal or length weights make the trajectory objective less faithful to high-quality candidate discovery.

\paragraph{Turn budget.}
Increasing $T_{\max}$ from one to three turns yields a clear gain, confirming that user intents benefit from multi-turn correction. The performance gains are substantial in the first three turns. Considering both efficiency and performance, we therefore use $T_{\max}=3$ in the main experiments to balance quality and inference cost.

\section{Controlled Sampling-Budget Comparison}
\label{app:sampling_budget_comparison}

\begin{figure}[h]
  \centering
  \includegraphics[width=0.92\linewidth]{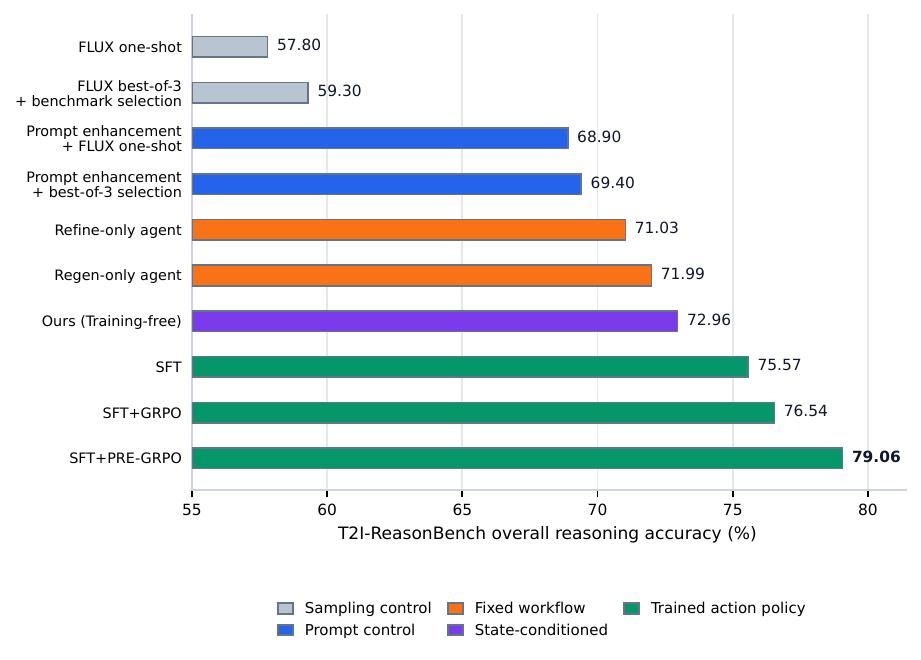}
  \caption{\textbf{Controlled sampling-budget comparison on T2I-ReasonBench.} We compare one-shot generation, best-of-3 controls with benchmark selection, prompt-enhanced controls, fixed-workflow agents, training-free state-conditioned action making, and trained action-policy variants. The small gains from best-of-3 selection isolate the effect of additional sampling, while the larger gains from state-conditioned and trained action policies show the net contribution of action-policy learning.}
  \label{fig:app_sampling_budget_comparison}
\end{figure}

We further analyze the performance gains brought by each individual operation. Fig.~\ref{fig:app_sampling_budget_comparison} separates the gain from extra sampling budget, prompt enhancement, fixed-workflow multi-turn execution, state-conditioned action making, and policy optimization on T2I-ReasonBench. The best-of-3 controls with benchmark selection improve only modestly over their corresponding one-shot settings: FLUX improves from 57.8 to 59.3, while prompt enhancement improves from 68.9 to 69.4. In contrast, fixed-workflow agents, training-free state-conditioned action making, and learned action-policy variants produce progressively larger gains, with \pregrpo{} reaching 79.06. This controlled view indicates that the final improvement is not explained by sampling budget alone; the main gains come from state-conditioned action making and trajectory-level training.

\section{Best-Score vs. Final-Score Selection}
\label{app:best_vs_final_selection}

\begin{figure}[h]
  \centering
  \includegraphics[width=0.55\linewidth]{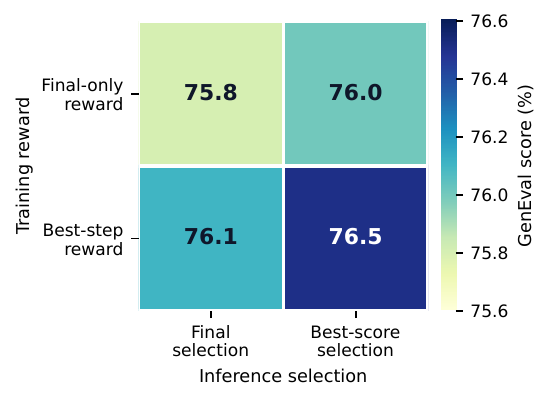}
  \caption{\textbf{Best-score vs. final-score selection on T2I-ReasonBench.} We ablate the training reward (final-step only vs. best-step) and the inference selector (final output vs. best-scored trajectory candidate). Best-score selection consistently improves over final-output selection under both reward choices.}
  \label{fig:app_best_vs_final_selection}
\end{figure}

In the main paper, we use the highest-scoring image along the generation trajectory as the delivered output rather than simply returning the final generated image. To examine this output rule, we compare final-output selection with best-score selection at inference. We also include two reward variants during training, a final-step-only reward and a best-step reward, to check whether the same selection trend holds under different training signals. Results on T2I-ReasonBench are shown in Fig.~\ref{fig:app_best_vs_final_selection}.

Across both reward variants, best-score selection outperforms final-output selection. This indicates that, in a multi-turn generation trajectory, the last image is not always the strongest candidate according to the evaluation score, so preserving the best-scored candidate is a simple and more reliable output rule. The combination of best-step reward and best-score selection yields the strongest result (76.5\%). We therefore adopt best-score trajectory selection as the default output strategy, including in the final \pregrpo{} setting.

\section{Average Generation Turns on GenEval}
\label{app:iteration_steps}

To further examine the simple-prompt behavior discussed in Section~\ref{sec:action_training_ablation}, we report the average number of generation turns consumed by each method on the GenEval benchmark (Table~\ref{tab:avg_steps}). This statistic is not meant to explain performance by turn count alone, since our final output selector keeps the highest-scored candidate across the trajectory. Instead, it serves as an indirect signal of how often each method proposes additional actions on prompts where the initial candidate is often already competitive.

\begin{table}[h]
\centering
\caption{Average generation turns on GenEval. PRE-GRPO uses fewer turns while achieving the best score, indicating better-calibrated action choices on simple compositional tasks.}
\label{tab:avg_steps}
\begin{tabular}{lcc}
\toprule
Method & Avg.\ Turns & GenEval score \\
\midrule
TF Agent        & 1.95 & 0.831 \\
SFT             & 1.97 & 0.826 \\
SFT + GRPO      & 1.89 & 0.854 \\
SFT + PRE-GRPO  & \textbf{1.67} & \textbf{0.884} \\
\bottomrule
\end{tabular}
\end{table}

GenEval prompts are compositionally simple (e.g., ``a red apple and a green bench''), meaning that a competent generator can often produce an adequate candidate with few turns. In this regime, the main risk is not merely using more turns, but choosing worse and unnecessary actions may hurt the candidate pool. A training-free agent can be overly sensitive to this mistake, while SFT imitates demonstrated actions without trajectory-level reward feedback. 

Table~\ref{tab:avg_steps} is consistent with this interpretation. TF Agent and SFT consume nearly two full turns on average ($1.95$ and $1.97$), yet both underperform the one-shot GenEval baseline reported in Fig.~\ref{fig:main_ablation}. Vanilla GRPO reduces the average only slightly ($1.89$) and recovers part of the score. PRE-GRPO lowers the average to $1.67$ while achieving the best GenEval score ($0.8843$), suggesting that its trajectory-level objective discourages low-value actions and yields better-calibrated action choices on simple prompts. 

\section{Effect of Turn Budget on Quality and Latency}
\label{app:latency_tradeoff}

\begin{figure}[H]
\centering
\includegraphics[width=0.55\linewidth]{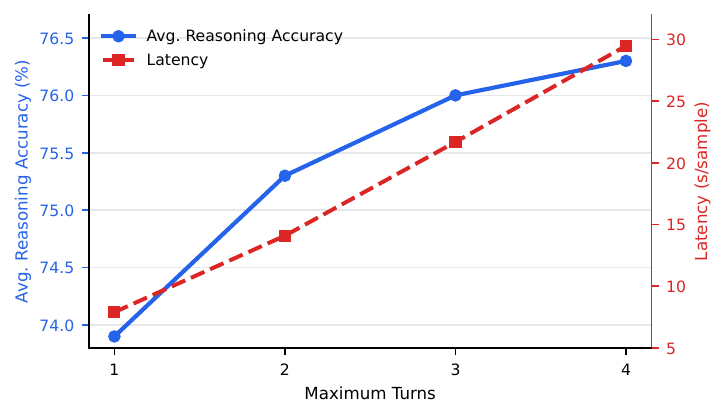}
\caption{Quality–latency trade-off as the turn budget increases from 1 to 4.
Users can adjust the maximum turn adaptively according to their computational budget.}
\label{fig:latency_tradeoff}
\end{figure}

We measure the quality-latency trade-off under different maximum turns. We vary the maximum turn budget from 1 to 4 using the no-CoT Navigator with
Qwen3-VL-8B-Instruct as the reviewer and evaluate on T2I-ReasonBench. Results
are visualized in Fig.~\ref{fig:latency_tradeoff}.

Latency increases linearly. Each additional turn introduces one extra generator inference and one additional critic model invocation, making the cost variation highly predictable. When the turn number \(T\) is set to 3, the system can retain most of the benefits brought by multi-turn interaction while keeping the latency within a reasonable range. Therefore, setting \(T=3\) as the default number of turns is a reasonable choice. Meanwhile, even when $T$ is small, the method already reaches performance well above
the one-shot FLUX.2-9B baseline. This means users can achieve satisfactory performance at a lower cost in practical applications.

\section{Latency Analysis on WISE}
\label{app:wise_latency}

\begin{figure}[h]
  \centering
  \includegraphics[width=0.58\linewidth]{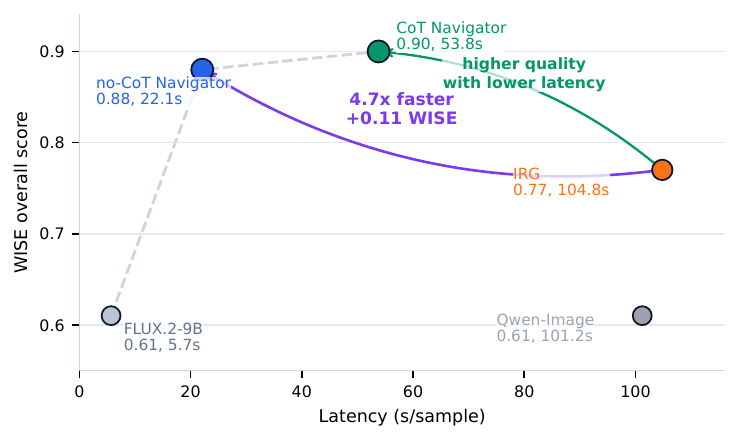}
  \caption{\textbf{Quality--latency trade-off on WISE.} Generation Navigator achieves substantially higher WISE scores than IRG while using much lower latency, especially in the no-CoT setting.}
  \label{fig:wise_latency_tradeoff}
\end{figure}

Complementing the quality--latency analysis in Section~\ref{sec:efficiency_human_validation}, we further compare different methods on the WISE benchmark~\citep{niu2025wise}. Fig.~\ref{fig:wise_latency_tradeoff} plots each method by its WISE overall score and per-sample latency. The base model FLUX.2-9B~\citep{blackforestlabs2026flux2klein} finishes in 5.7 seconds but only reaches a WISE score of 0.61. Qwen-Image~\citep{wu2025qwenimage} obtains the same WISE score of 0.61 while requiring 101.2 seconds. The agent-based method IRG~\citep{huang2026interleaving} improves quality to 0.77, but its latency increases to 104.8 seconds, about 18.4$\times$ slower than FLUX.2-9B, due to complex LLM reasoning and full-image regeneration within its closed loop.

Generation Navigator achieves a WISE score of 0.90 with CoT reasoning at 53.8 seconds. Removing CoT reduces the latency to 22.1 seconds while retaining a high WISE score of 0.88. Compared with IRG, the no-CoT Navigator is 4.7$\times$ faster and improves the WISE score by 0.11. This shows that Generation Navigator provides a stronger quality--latency trade-off than existing agent methods on knowledge-intensive generation.

\clearpage
\section{Qualitative Visualizations}
\label{app:qualitative_visualizations}

We provide representative qualitative cases covering textual image design, counting, spatial-relation, and two-object prompts. The top row shows the one-shot baseline and the navigator trajectory as vector-labeled image cards. The baseline is the generator output from the original prompt. For navigator images, labels indicate the action that produced the image, followed by the reviewer score and next action.

\subsection{Case 1: Textual Image Design}

\noindent\textbf{Original prompt.} Design a colorful book cover featuring a caricature of Mozart playing the piano.

\begin{center}
\qualbasecard{0.30\linewidth}{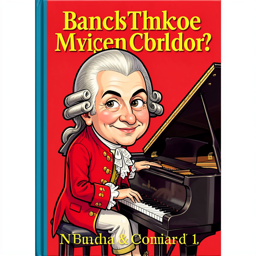}{One-shot baseline}{generator only}
\hfill
\qualcard{0.30\linewidth}{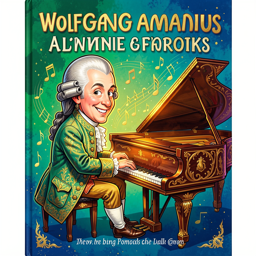}{Turn 1: initial T2I}{score 3.30\\next: \regenact}
\hfill
\qualcard{0.30\linewidth}{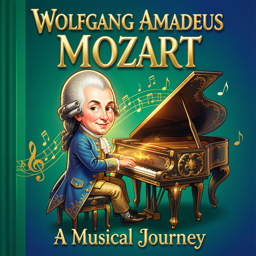}{Turn 2: by \regenact}{score 4.88\\next: \stopact}
\end{center}

\paragraph{Navigator turn details.}
\begin{tcolorbox}[qualdetail,title={Turn 1: initial T2I; next action \regenact{} (reviewer score 3.30)}]
\footnotesize
\textbf{Generation prompt.} A vibrant, colorful book cover features a whimsical caricature of \qualkey{Wolfgang Amadeus Mozart} playing a grand piano. The cover bursts with bold, saturated hues—emerald, sapphire, and gold—creating a lively backdrop. Mozart, in a playful caricature style, wears an exaggerated powdered wig, a detailed 18th - century coat with intricate patterns, and plays a grand piano with ornate carvings and glowing keys. Musical notes swirl around him, enhancing the musical theme. The design blends classical artistry with humorous exaggeration, ideal for a music - themed book, capturing Mozart’s genius in a fun, eye - catching, artistically rich composition.

\textbf{Reviewer diagnosis.} All text content on this book cover is \qualkey{nonsensical AI-generated gibberish}. Mozart's name is incorrectly rendered as \qualkey{garbled nonsense text} instead of the correct name, and the small text at the bottom of the cover is also meaningless unreadable garbage, which fails to meet the requirement of a proper functional book cover. The core visual elements (colorful style, caricature of Mozart playing piano) are correctly implemented, the overall visual aesthetic is good with no major technical defects.
\end{tcolorbox}

\begin{tcolorbox}[qualdetail,title={Turn 2: generated by \regenact{}; next action \stopact{} (reviewer score 4.88)}]
\footnotesize
\textbf{Generation prompt.} Vibrant, colorful book cover (emerald, sapphire, gold hues) features a whimsical caricature of \qualkey{Wolfgang Amadeus Mozart} (exaggerated powdered wig, detailed 18th - century coat) playing a grand piano (ornate carvings, glowing keys). Musical notes swirl around. \qualkey{Top text: “WOLFGANG AMADEUS MOZART”} in bold golden 18th - century - inspired font. \qualkey{Bottom text: “A Musical Journey” (legible)} in 18th - century - style. Design blends classical artistry with humorous exaggeration, ideal for a music - themed book, capturing Mozart’s genius in a fun, eye - catching, artistically rich composition.

\textbf{Reviewer diagnosis.} There are \qualkey{no significant flaws}. The image fully meets all specified requirements: it is a properly formatted colorful book cover, correctly features a clear caricature of \qualkey{Wolfgang Amadeus Mozart} playing a grand piano, with high technical quality, harmonious color matching, good layered lighting, sharp clear details, and cohesive subject-background integration. 
\end{tcolorbox}

\clearpage
\subsection{Case 2: Spatial Relation}

\noindent\textbf{Original prompt.} a photo of a bus above a boat.

\begin{center}
\qualbasecard{0.225\linewidth}{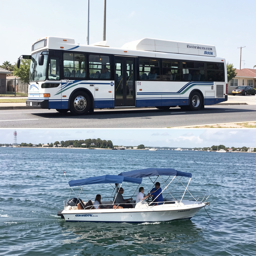}{One-shot baseline}{generator only}
\hfill
\qualcard{0.225\linewidth}{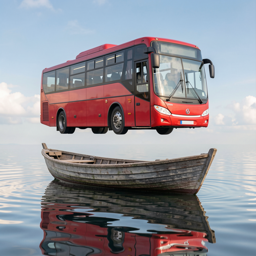}{Turn 1: initial T2I}{score 3.30\\next: \regenact}
\hfill
\qualcard{0.225\linewidth}{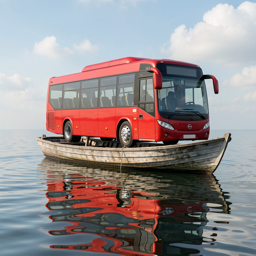}{Turn 2: by \regenact}{score 4.82\\next: \regenact}
\hfill
\qualcard{0.225\linewidth}{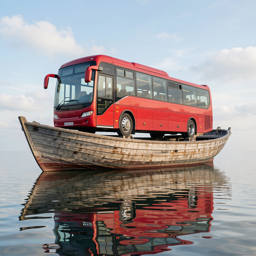}{Turn 3: by \regenact}{score 4.85\\next: \stopact}
\end{center}

\paragraph{Navigator turn details.}
\begin{tcolorbox}[qualdetail,title={Turn 1: initial T2I; next action \regenact{} (reviewer score 3.30)}]
\footnotesize
\textbf{Generation prompt.} A photorealistic image depicts a vibrant red city bus (with large windows, sleek modern design) \qualkey{floating above} a small, weathered wooden rowboat. The bus \qualkey{hovers mid - air}, positioned directly above the boat, which rests on calm, rippled water that reflects the bus’s underside. The background features a clear sky with soft clouds, enhancing the surreal, dreamlike atmosphere. The scene contrasts the bus’s sleek design with the boat’s rustic texture, ideal for artistic photography, capturing the unexpected bus - over - boat juxtaposition.

\textbf{Reviewer diagnosis.} The bus is \qualkey{unrealistically levitating in mid-air} above the empty boat, rather than being \qualkey{supported on the boat} as expected for a bus above a boat, creating a physically impossible, unnatural arrangement that does not match the implicit expectation of a plausible photo.
\end{tcolorbox}

\begin{tcolorbox}[qualdetail,title={Turn 2: generated by \regenact{}; next action \regenact{} (reviewer score 4.82)}]
\footnotesize
\textbf{Generation prompt.} A photorealistic image depicts a vibrant red city bus (sleek modern design, large windows) \qualkey{resting on top of a small, weathered wooden rowboat}. The bus is positioned directly atop the boat, which floats on calm, rippled water that reflects the bus’s underside and the boat’s weathered surface. The background features a clear sky with soft clouds, enhancing the surreal, dreamlike atmosphere. The scene contrasts the bus’s sleek design with the boat’s rustic texture, ideal for artistic photography, capturing the \qualkey{bus - supported - by - boat juxtaposition (not floating mid - air)}.

\textbf{Reviewer diagnosis.} The image correctly meets the core requirement of positioning a bus above a boat, with high overall visual quality. The only flaw is that the \qualkey{small wooden rowboat is physically implausible} to support the \qualkey{large size and weight of the full-sized passenger bus}, which violates real-world physical consistency.
\end{tcolorbox}

\begin{tcolorbox}[qualdetail,title={Turn 3: generated by \regenact{}; next action \stopact{} (reviewer score 4.85)}]
\footnotesize
\textbf{Generation prompt.} A photorealistic image depicts a vibrant red city bus (sleek modern design, large windows) resting on a \qualkey{large, sturdy, weathered wooden boat (replacing the small rowboat)} that can \qualkey{support the bus’s weight}. The bus is positioned atop the larger, robust vessel, which floats on calm, rippled water reflecting the bus’s underside and the boat’s weathered surface. The background features a clear sky with soft clouds, enhancing the surreal, dreamlike atmosphere. The scene contrasts the bus’s sleek design with the boat’s rugged, sturdy texture, ideal for artistic photography, capturing the \qualkey{bus - supported - by - larger - boat juxtaposition (not floating mid - air)}.

\textbf{Reviewer diagnosis.} The image \qualkey{fully meets the user's requirement} of showing a bus positioned above a boat, with \qualkey{no critical flaws}. It has natural lighting, accurate color reproduction, clear subject details, and harmonious composition with the background. There are only very minor shortcomings: it lacks ultra-high resolution micro-texture details, which does not affect the overall presentation and compliance with the request.

\end{tcolorbox}

\clearpage
\subsection{Case 3: Spatial Relation with Refinement}

\noindent\textbf{Original prompt.} a photo of a frisbee right of a motorcycle.

\begin{center}
\qualbasecard{0.225\linewidth}{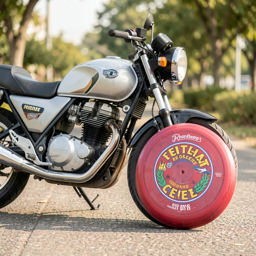}{One-shot baseline}{generator only}
\hfill
\qualcard{0.225\linewidth}{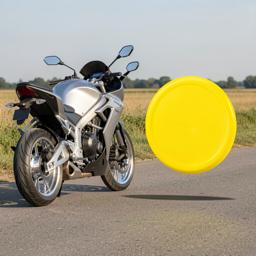}{Turn 1: initial T2I}{score 3.40\\next: \refineact}
\hfill
\qualcard{0.225\linewidth}{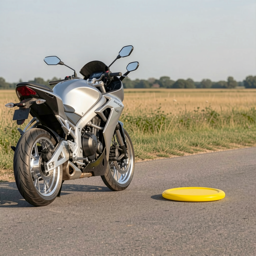}{Turn 2: by \refineact}{score 4.70\\next: \regenact}
\hfill
\qualcard{0.225\linewidth}{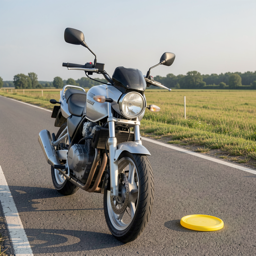}{Turn 3: by \regenact}{score 4.79\\next: \stopact}
\end{center}

\paragraph{Navigator turn details.}
\begin{tcolorbox}[qualdetail,title={Turn 1: initial T2I; next action \refineact{} (reviewer score 3.40)}]
\footnotesize
\textbf{Generation prompt.} A photorealistic photo depicts a sleek motorcycle (metallic silver, black accents, chrome wheels) on the left, with a vibrant circular frisbee (bright yellow, smooth surface) \qualkey{positioned to its right}. The backdrop is a neutral outdoor setting (e.g., paved road or grassy field), lit by natural sunlight. The motorcycle is stationary, while the frisbee \qualkey{rests or floats}, creating a clear left - right spatial arrangement. Details include the motorcycle’s streamlined body, handlebars, and the frisbee’s glossy texture, rendered with sharp, realistic detail for a high - quality image.

\textbf{Reviewer diagnosis.} While the frisbee is correctly positioned to the right of the motorcycle as requested, the frisbee is \qualkey{unnaturally floating in mid-air} above the road instead of being placed naturally in the scene. The frisbee is \qualkey{obviously artificially added, poorly integrated with the environment}, and looks inconsistent with the realistic style of the background and motorcycle, creating a jarring visual effect.
\end{tcolorbox}

\begin{tcolorbox}[qualdetail,title={Turn 2: generated by \refineact{}; next action \regenact{} (reviewer score 4.70)}]
\footnotesize
\textbf{Generation prompt.} Make the \qualkey{bright yellow frisbee lie on the paved road to the right of the motorcycle}, ensuring it appears \qualkey{naturally placed and integrated} with the road surface and the outdoor setting.

\textbf{Reviewer diagnosis.} The image correctly fulfills the core request: the frisbee is placed \qualkey{to the right of the motorcycle}, and both required objects are present. There are minor AI generation flaws: the motorcycle incorrectly has \qualkey{three side mirrors} instead of the standard two, and the model text on the motorcycle body is \qualkey{garbled unreadable gibberish}. The rest of the image has good technical quality, natural lighting and clear details.
\end{tcolorbox}

\begin{tcolorbox}[qualdetail,title={Turn 3: generated by \regenact{}; next action \stopact{} (reviewer score 4.79)}]
\footnotesize
\textbf{Generation prompt.} A photo of a silver motorcycle with \qualkey{two standard side mirrors} and \qualkey{clear legible model text}, parked on a paved road. A \qualkey{bright yellow frisbee lies naturally on the road to the right of the motorcycle}. The background features a grassy field with distant trees under a clear sky, with natural lighting and sharp details. Photorealistic style, ensuring the motorcycle’s mirror count and text are accurate.

\textbf{Reviewer diagnosis.} The image correctly meets the requirement: the \qualkey{yellow frisbee is placed to the right of the motorcycle}, as requested. Only a minor subtle flaw exists: the \qualkey{text branding on the motorcycle's body is garbled/distorted}, which is a small AI generation artifact. The image has good overall technical quality, clear details, harmonious natural colors and proper lighting.

\end{tcolorbox}

\clearpage
\subsection{Case 4: Counting with Refinement and Regeneration}

\noindent\textbf{Original prompt.} a photo of three benches.

\begin{center}
\qualbasecard{0.225\linewidth}{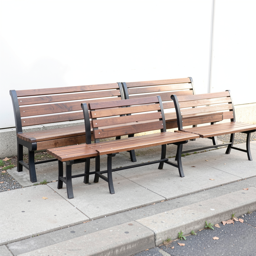}{One-shot baseline}{generator only}
\hfill
\qualcard{0.225\linewidth}{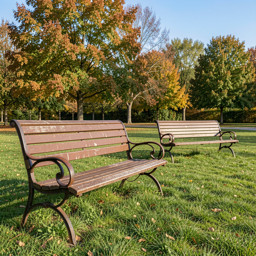}{Turn 1: initial T2I}{score 2.60\\next: \refineact}
\hfill
\qualcard{0.225\linewidth}{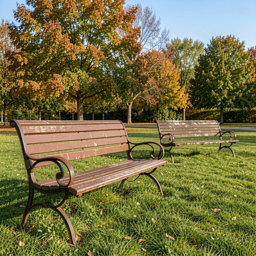}{Turn 2: by \refineact}{score 2.69\\next: \regenact}
\hfill
\qualcard{0.225\linewidth}{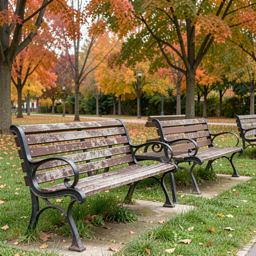}{Turn 3: by \regenact}{score 4.85\\selected at budget end}
\end{center}

\paragraph{Navigator turn details.}
\begin{tcolorbox}[qualdetail,title={Turn 1: initial T2I; next action \refineact{} (reviewer score 2.60)}]
\footnotesize
\textbf{Generation prompt.} \qualkey{Three weathered brown wooden benches} with metal armrests and legs are arranged in a park - like setting: \qualkey{one in the foreground, one in the mid - ground, and one in the background}, creating a sense of depth. Lush green grass, trees with autumnal foliage, and a clear blue sky form the backdrop. Each bench has a slatted seat and backrest, with subtle wood wear and slightly rusted metal, suggesting regular use. Styled as a high - resolution, naturalistic photograph, it is ideal for showcasing outdoor furniture in a peaceful environment.

\textbf{Reviewer diagnosis.} The user requested a photo of \qualkey{three benches}, but the current image only contains \qualkey{two park benches}, failing to meet the required quantity of benches, which is the key flaw. The image itself has good overall visual quality, with no obvious technical or aesthetic defects.
\end{tcolorbox}

\begin{tcolorbox}[qualdetail,title={Turn 2: generated by \refineact{}; next action \regenact{} (reviewer score 2.69)}]
\footnotesize
\textbf{Generation prompt.} Add a \qualkey{third weathered brown wooden bench} with metal armrests and legs \qualkey{in the background}, positioned behind the mid - ground bench to maintain the depth arrangement, matching the style and weathered appearance of the existing two benches.

\textbf{Reviewer diagnosis.} The image only contains \qualkey{2 park benches}; it fails to meet the user's explicit requirement of \qualkey{three benches}, with the \qualkey{third bench completely missing} from the scene. There are no other significant technical or aesthetic flaws to the image itself.
\end{tcolorbox}

\begin{tcolorbox}[qualdetail,title={Turn 3: generated by \regenact{}; selected at budget end (reviewer score 4.85)}]
\footnotesize
\textbf{Generation prompt.} Add a \qualkey{third weathered brown wooden park bench} with metal armrests and legs, positioned \qualkey{behind the mid - ground bench (the right - hand bench)} to form a \qualkey{three - bench depth arrangement}, matching the weathered appearance and style of the existing two benches, set in a park with autumn - hued trees and green grass, under natural daylight.

\textbf{Reviewer diagnosis.} The \qualkey{rightmost (third) bench is partially cropped} by the image boundary, with only half of its structure visible in the frame. There are no other significant technical, color, lighting, or composition flaws; the image is otherwise sharp, well-exposed, and harmoniously composed.

\textbf{Termination.} The next suggested action was a \refineact{} to fix cropping, but the turn budget was exhausted; the output selector keeps this best-scored candidate.
\end{tcolorbox}

\clearpage
\subsection{Case 5: Scientific Reasoning}

\noindent\textbf{Original prompt.} A cup of icy coffee in warm humid air.

\begin{center}
\qualbasecard{0.225\linewidth}{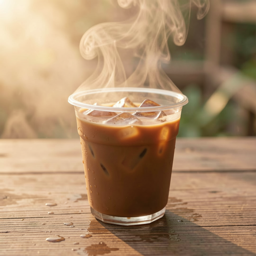}{One-shot baseline}{generator only}
\hfill
\qualcard{0.225\linewidth}{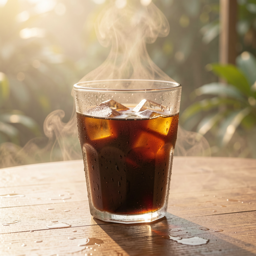}{Turn 1: initial T2I}{score 3.10\\next: \regenact}
\hfill
\qualcard{0.225\linewidth}{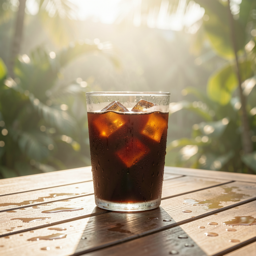}{Turn 2: by \regenact}{score 4.74\\next: \refineact}
\hfill
\qualcard{0.225\linewidth}{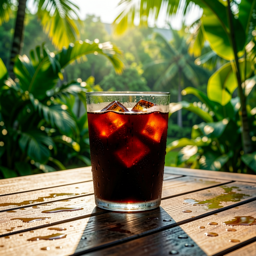}{Turn 3: by \refineact}{score 4.91\\next: \stopact}
\end{center}

\paragraph{Navigator turn details.}
\begin{tcolorbox}[qualdetail,title={Turn 1: initial T2I; next action \regenact{} (reviewer score 3.10)}]
\footnotesize
\textbf{Generation prompt.} A clear glass cup of iced coffee---filled with dark coffee and ice cubes---rests in a warm, humid atmosphere. Condensation droplets glisten on the cup's exterior, formed as the cool drink interacts with the moist, sunlit air. The air around the cup is heavy with humidity, creating a hazy, tropical ambiance (e.g., a sunlit patio) to emphasize the heat. The scene, rendered in a photorealistic style, captures the contrast between the cold, refreshing coffee and the warm, humid surroundings, with detailed condensation and atmospheric moisture.

\textbf{Reviewer diagnosis.} The image has a core content error that conflicts with the request: \qualkey{visible rising steam is incorrectly added} to the iced coffee. Cold icy coffee \qualkey{cannot produce visible steam} in warm humid air; steam is a characteristic of hot beverages, not iced coffee. While the condensation droplets on the glass correctly reflect the humid environment, the erroneous steam directly contradicts the icy-coffee requirement.
\end{tcolorbox}

\begin{tcolorbox}[qualdetail,title={Turn 2: generated by \regenact{}; next action \refineact{} (reviewer score 4.74)}]
\footnotesize
\textbf{Generation prompt.} A clear glass cup of iced coffee---filled with dark coffee and ice cubes---rests on a sunlit wooden patio table in warm, humid air. Condensation droplets glisten on the cup's exterior, formed as the cool drink interacts with the moist, sunlit air. The air around the cup is heavy with humidity, creating a hazy, tropical ambiance (lush green foliage in the background) to emphasize the heat. The scene, rendered in a photorealistic style, captures the contrast between the cold, refreshing coffee and the warm, humid surroundings, with detailed condensation and atmospheric moisture---\qualkey{no visible steam} (since iced coffee does not produce steam in warm air). The table shows water droplets from condensation, and the background features sunlit greenery to enhance the tropical, humid setting.

\textbf{Reviewer diagnosis.} There are \qualkey{no significant flaws} that violate the request. Only minor softening of the out-of-focus background foliage remains; the image accurately presents \qualkey{icy coffee in warm humid air}, with proper condensation on the glass that demonstrates the physical effect of a cold beverage in a humid warm environment.
\end{tcolorbox}

\begin{tcolorbox}[qualdetail,title={Turn 3: generated by \refineact{}; next action \stopact{} (reviewer score 4.91)}]
\footnotesize
\textbf{Generation prompt.} Make the background foliage \qualkey{more sharply defined and vibrant} to reduce the softening effect, enhancing the tropical, humid ambiance.

\textbf{Reviewer diagnosis.} There are \qualkey{no significant flaws}. Only a minor imperfection remains: a small number of liquid stains on the wooden table look slightly artificial, but this does not affect the overall presentation. The image correctly conveys the effect of \qualkey{iced coffee in warm humid air}, with accurate condensation on the glass and an appropriate warm natural environment.

\end{tcolorbox}

\clearpage
\subsection{Case 6: Spatial Relation}

\noindent\textbf{Original prompt.} a photo of a book above a laptop.

\begin{center}
\qualbasecard{0.225\linewidth}{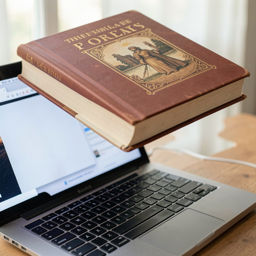}{One-shot baseline}{generator only}
\hfill
\qualcard{0.225\linewidth}{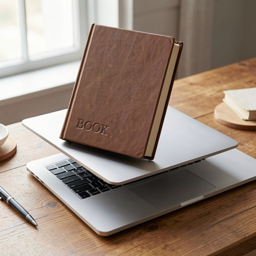}{Turn 1: initial T2I}{score 2.75\\next: \regenact}
\hfill
\qualcard{0.225\linewidth}{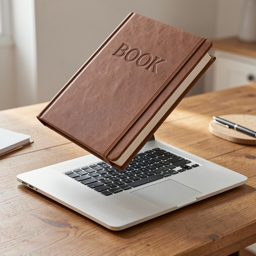}{Turn 2: by \regenact}{score 4.41\\next: \regenact}
\hfill
\qualcard{0.225\linewidth}{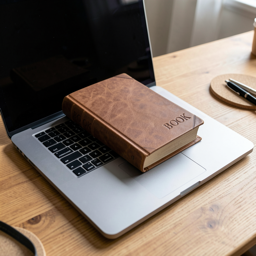}{Turn 3: by \regenact}{score 4.91\\next: \stopact}
\end{center}

\paragraph{Navigator turn details.}
\begin{tcolorbox}[qualdetail,title={Turn 1: initial T2I; next action \regenact{} (reviewer score 2.75)}]
\footnotesize
\textbf{Generation prompt.} A photorealistic image depicts a brown hardcover book positioned vertically above a sleek silver laptop, creating a stacked arrangement. The laptop, open with its screen facing upward, rests on a wooden desk, its keyboard and trackpad visible. The book, with a textured cover and ``BOOK'' embossed on the spine, sits atop the laptop’s lid, slightly tilted to suggest balance. The desk features subtle details: a pen and a coaster, adding realism. Natural lighting highlights the book’s texture and the laptop’s metallic sheen, rendered in a detailed, photorealistic style to emphasize the spatial relationship between the two objects.

\textbf{Reviewer diagnosis.} The image incorrectly includes \qualkey{two stacked laptops} (an open base laptop with an extra closed laptop stacked on top of it) instead of the requested \qualkey{single laptop with the book positioned above it}, which deviates from the original user requirement. The image itself has good overall aesthetic and technical quality, only the core element composition does not match the request.
\end{tcolorbox}

\begin{tcolorbox}[qualdetail,title={Turn 2: generated by \regenact{}; next action \regenact{} (reviewer score 4.41)}]
\footnotesize
\textbf{Generation prompt.} A photorealistic image depicts a brown hardcover book positioned vertically above a \qualkey{single sleek silver laptop}, with the laptop open (screen facing upward) on a wooden desk. The laptop, with its keyboard and trackpad visible, rests on the desk. The book, featuring a textured cover and ``BOOK'' embossed on the spine, sits atop the laptop’s lid, slightly tilted to suggest balance. The desk includes a pen and a coaster, adding realism. Natural lighting highlights the book’s texture and the laptop’s metallic sheen, rendered in a detailed, photorealistic style to emphasize the spatial relationship between the book and the \qualkey{single laptop}.

\textbf{Reviewer diagnosis.} The book is \qualkey{unrealistically levitating/floating unsupported} above the laptop keyboard, creating a physically impossible, unnatural arrangement which is an AI generation artifact. Aside from this issue, the image has good overall technical quality, harmonious colors and clear details.
\end{tcolorbox}

\begin{tcolorbox}[qualdetail,title={Turn 3: generated by \regenact{}; next action \stopact{} (reviewer score 4.91)}]
\footnotesize
\textbf{Generation prompt.} A photorealistic image depicts a brown hardcover book \qualkey{resting (not floating) on the lid} of a \qualkey{single sleek silver laptop}, with the laptop open (screen facing upward) on a wooden desk. The laptop, with its keyboard and trackpad visible, rests on the desk. The book, featuring a textured cover and ``BOOK'' embossed on the spine, sits atop the laptop’s lid \qualkey{balanced naturally, no floating} on the desk. The desk includes a pen and a coaster, adding realism. Natural lighting highlights the book’s texture and the laptop’s metallic sheen, rendered in a detailed, photorealistic style to emphasize the correct spatial relationship between the book and the single laptop.

\textbf{Reviewer diagnosis.} \qualkey{No significant flaws}. The image correctly meets all requirements: it shows a \qualkey{book placed above a laptop} as requested, has clear sharp details, natural soft lighting, harmonious natural colors, and good integration between the subject and the background.
\end{tcolorbox}

\clearpage
\subsection{Case 7: Two-Object Repair}

\noindent\textbf{Original prompt.} a photo of a toothbrush and a bench.

\begin{center}
\qualbasecard{0.225\linewidth}{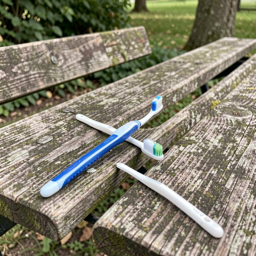}{One-shot baseline}{generator only}
\hfill
\qualcard{0.225\linewidth}{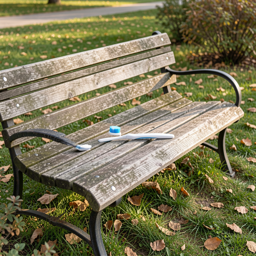}{Turn 1: initial T2I}{score 3.30\\next: \regenact}
\hfill
\qualcard{0.225\linewidth}{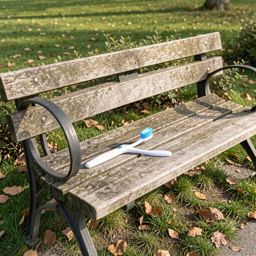}{Turn 2: by \regenact}{score 3.30\\next: \regenact}
\hfill
\qualcard{0.225\linewidth}{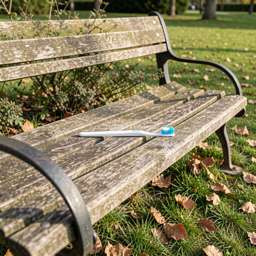}{Turn 3: by \regenact}{score 4.85\\next: \stopact}
\end{center}

\paragraph{Navigator turn details.}
\begin{tcolorbox}[qualdetail,title={Turn 1: initial T2I; next action \regenact{} (reviewer score 3.30)}]
\footnotesize
\textbf{Generation prompt.} A photorealistic image features a toothbrush and a bench in a serene outdoor park setting. The bench, crafted from weathered wooden planks with metal armrests, rests on a grassy lawn dotted with fallen leaves and small shrubs. The toothbrush, with a white handle and blue bristles, lies on the bench’s seat, positioned centrally, its bristles slightly angled. Soft natural light casts gentle shadows, highlighting the contrast between the everyday toothbrush and the rustic, weathered bench, creating a calm, everyday-scene aesthetic.

\textbf{Reviewer diagnosis.} The request asks for \qualkey{one toothbrush and a bench}, but this image contains \qualkey{two toothbrushes}. One toothbrush is incorrectly generated, being \qualkey{abnormally embedded into the left metal armrest} of the bench, which is physically implausible and is a generation artifact.
\end{tcolorbox}

\begin{tcolorbox}[qualdetail,title={Turn 2: generated by \regenact{}; next action \regenact{} (reviewer score 3.30)}]
\footnotesize
\textbf{Generation prompt.} A photorealistic image features a \qualkey{single toothbrush} and a bench in a serene outdoor park setting. The bench, crafted from weathered wooden planks with metal armrests, rests on a grassy lawn dotted with fallen leaves and small shrubs. The toothbrush, with a white handle and blue bristles, lies on the bench’s seat \qualkey{not embedded in the armrest}, positioned centrally, its bristles slightly angled. Soft natural light casts gentle shadows, highlighting the contrast between the everyday toothbrush and the rustic, weathered bench, creating a calm, everyday-scene aesthetic.

\textbf{Reviewer diagnosis.} The toothbrush is incorrectly generated as a \qualkey{distorted, impossible malformed object}: it has a \qualkey{single brush head connected to two separate full handles}, which does not match the normal structure of a standard toothbrush requested. The bench is correctly presented, the background and overall image technical quality are good.
\end{tcolorbox}

\begin{tcolorbox}[qualdetail,title={Turn 3: generated by \regenact{}; next action \stopact{} (reviewer score 4.85)}]
\footnotesize
\textbf{Generation prompt.} A photorealistic image features a \qualkey{standard toothbrush (single white handle, single blue-bristled brush head)} and a weathered wooden bench with metal armrests in a serene outdoor park. The bench rests on a grassy lawn dotted with fallen leaves and small shrubs. The toothbrush lies centrally on the bench’s seat \qualkey{not embedded in the armrest}, its bristles slightly angled. Soft natural light casts gentle shadows, highlighting the contrast between the everyday toothbrush and the rustic, weathered bench, creating a calm, everyday-scene aesthetic.

\textbf{Reviewer diagnosis.} Both requested subjects (\qualkey{a toothbrush and a park bench}) are present and clearly visible in the image. There are \qualkey{no major technical or content flaws}: the image has sharp, clear details, natural accurate exposure, harmonious natural outdoor coloring, no significant artifacts, noise or distortion. All requirements of the request are fully met.
\end{tcolorbox}

\clearpage
\section{Human Evaluation}
\label{app:human_evaluation}

We conduct a pairwise human evaluation as an empirical validation of the automatic reviewer used throughout training and evaluation. The goal is not to replace benchmark evaluation, but to test whether reviewer-induced preferences are reasonably aligned with human judgments on generated images.

\paragraph{Sample construction.}
We sample some image pairs from two pools: T2I-ReasonBench and GenEval results. For each image, the reviewer produces a scalar score
\[
\rho = 0.3 \cdot \text{Visual Quality} + 0.7 \cdot \text{Instruction Comprehension}.
\]
Pairwise reviewer preferences are derived from score differences: a pair is treated as a reviewer tie when $|\rho_A-\rho_B|<0.3$, and otherwise the image with the higher $\rho$ is preferred. To cover both easy and ambiguous cases, pairs are sampled across four score-gap bins: tie ($<0.3$), small ($0.3$--$0.8$), medium ($0.8$--$1.6$), and large ($\geq1.6$).

\paragraph{Annotation protocol.}
Annotators are shown the original prompt and two anonymized candidate images in randomized left-right order. Source run information, turn identifiers, and reviewer scores are hidden. When available, prompts are displayed in both English and Chinese. Each annotator selects whether the left image is better, the right image is better, or the two images are tied. The interface assigns each annotator a deterministic random subset of 40 pairs.

\paragraph{Metric and result.}
Agreement is computed by comparing the human pairwise preference with the reviewer-induced pairwise preference. In the current study, we collect 320 annotations from eight annotators. The MLLM reviewer's relative ranking agrees with human preference on approximately 70.3\% of decisive non-tie comparisons. Given the subjectivity of multimodal evaluation and the ambiguity of near-tie image pairs, this result suggests that the reviewer provides a usable trajectory-optimization signal, while still leaving room for richer human studies in future evaluations.

\section{Limitations and Future Directions}
\label{app:limitations_future}
Like all agentic workflows, Generation Navigator inherits two structural costs compared with single-shot generation: additional inference latency from iterative action making, and dependence on an external reviewer as the environment signal. We mitigate both within the current framework. For latency, the no-CoT navigator variant reduces per-turn action-making overhead by 2.4$\times$ while still achieving an 18.3-point accuracy improvement compared to the baseline (Section~\ref{sec:efficiency_human_validation}). The turn budget is also user-configurable; even with a small number of turns, the navigator substantially outperforms single-shot baselines (Appendix~\ref{app:latency_tradeoff}), providing a practical operating point that balances quality and cost. Meanwhile, compared with the recent agent-based method IRG~\citep{huang2026interleaving}, the no-CoT Navigator is 4.7$\times$ faster on WISE while improving the overall score from 0.77 to 0.88, suggesting that our closed-loop design offers a more favorable quality--latency trade-off than prior agentic workflows (Appendix~\ref{app:wise_latency}).
For reviewer dependence, the learned state-conditioned action policy is not tied to any single scorer: replacing the inference-time reviewer across five architecturally distinct models yields only a 3.03-point spread (Section~\ref{sec:robustness_transfer}), preserving a large gain over one-shot generation and confirming that the navigator acquires general visual-state action logic rather than reviewer-specific preferences. This decoupling also suggests a natural upgrade path: as foundational multimodal models continue to advance, stronger reviewers can be plugged in at inference time to provide richer feedback without retraining the navigator.

\section*{LLM Usage}
We acknowledge the use of a large language model (LLM) to assist in the preparation of this manuscript. The LLM's role was strictly limited to improving grammar and refining language. It did not contribute to any of the core research components, such as the initial ideas, experimental design, data analysis, or interpretation of the results.

\end{document}